\documentclass[10pt,a4paper]{article}
\usepackage[margin=2cm]{geometry}
\usepackage{amsmath,amssymb,amsthm}

\usepackage{booktabs}
\usepackage{graphicx}
\usepackage{hyperref}
\usepackage{url}
\usepackage{enumitem}
\usepackage{xcolor}
\usepackage{algorithm}
\usepackage{algpseudocode}
\usepackage{natbib}
\usepackage{caption}
\usepackage{multicol}
\usepackage{pgfplots}
\pgfplotsset{compat=1.18}
\usetikzlibrary{positioning,shapes.geometric}

\hypersetup{colorlinks=true, linkcolor=blue!60!black, citecolor=blue!60!black, urlcolor=blue!70!black}

\title{\textbf{The Alignment Tax: Response Homogenization in Aligned LLMs \\
and Its Implications for Uncertainty Estimation}}
\author{Mingyi Liu \\ Independent Researcher \\ GitHub: \texttt{@DigitLion}}
\date{March 25, 2026}

\begin{document}
\maketitle

\begin{abstract}
RLHF-aligned language models exhibit \textbf{response homogenization}: on TruthfulQA ($n$=790), 40\% of questions produce a single semantic cluster across 10 i.i.d.\ samples under Jaccard bigram clustering (79\% under embedding cosine clustering; robust across sample sizes $N$=3--10, temperatures $T$=0.3--1.5, and generation lengths 40--200 tokens; 33.5\% SCR persists at 200 tokens on 200 questions vs.\ 0\% for the base model). On affected questions, sampling-based uncertainty methods have zero discriminative power (AUROC=0.500), while free token entropy retains signal (0.603). This \textbf{alignment tax} is task-dependent: on GSM8K ($n$=500), token entropy achieves 0.724 (Cohen's $d$=0.81).

A base-vs-instruct ablation on Qwen3-14B confirms the causal role of alignment: the base model shows 1.0\% single-cluster rate vs.\ 28.5\% for the instruct model (Wilcoxon $p < 10^{-6}$). A three-way training stage ablation (Base 0.0\% $\to$ SFT 1.5\% $\to$ DPO 4.0\% SCR, all $p < 0.003$) localizes the cause to DPO, not SFT. Cross-family replication on four model families (Qwen3-14B: 28.5\%, LLaMA-3.2-3B: 5.5\%, Mistral-7B: 1.0\% SCR) reveals alignment tax severity varies by family and scale. Cross-chain replication on Tulu-3 (Llama-3.1-8B$\to$SFT$\to$DPO+RLVR) shows minimal alignment tax (0.5\% SCR vs.\ Zephyr's 4.0\%), confirming severity is recipe-dependent. Stage-wise token entropy measurement (Base 1.175$\to$SFT 1.055$\to$DPO 0.776 nats, $p$=1.17$\times 10^{-40}$) reveals the mechanistic \emph{decoupling}: DPO retains 66\% of token-level entropy while collapsing response diversity, explaining why B1 retains signal where sampling-based methods fail. We validate this finding across twenty-five experiments, five benchmarks, four model families, and three model scales (3B--14B), with proxy (Jaccard), SINdex-style embedding (agglomerative cosine clustering), cross-encoder NLI-SE head-to-head ($\Delta$AUROC=+0.014 vs.\ Jaccard, CIs overlap), and canonical NLI-based baselines at three DeBERTa scales (large 435M, base 184M, xsmall 70M---all $\approx$0.51 AUROC). Cross-embedder validation with two independent embedding families (Qwen3-Embedding 78\% SCR vs.\ Nomic-embed-text 92\% SCR at $\tau$=0.85) rules out coupling bias. Cross-dataset validation on WebQuestions (58.0\% SCR at $\tau$=0.85) confirms the alignment tax generalizes beyond TruthfulQA. LLM-judge labels are validated against TruthfulQA gold answer templates ($\kappa$=0.487). The central finding---response homogenization---is implementation-independent and label-free. Motivated by this diagnosis, we explore a cheapest-first cascade (\textbf{UCBD}) over orthogonal uncertainty signals. Selective prediction raises GSM8K accuracy from 84.4\% to 93.2\% at 50\% coverage; weakly dependent boundaries ($|r|\!\leq\!0.12$) enable 57\% cost savings.
\end{abstract}

\section{Introduction}

LLM-powered AI Agents are remarkably capable, yet one question has received surprisingly little systematic attention: \textbf{can an Agent recognize that it doesn't know something?}

Consider mathematical reasoning: on GSM8K ($n$=500), a 14B model's token entropy achieves AUROC=0.724 for detecting errors (Cohen's $d$=0.81), enabling selective prediction that raises accuracy from 84.4\% to 93.2\% at 50\% coverage. Yet the \emph{same} entropy signal on factual QA (TruthfulQA, overall) barely exceeds chance (0.52). This $10\times$ gap in effect size (Cohen's $d$: 0.81 vs.\ 0.07) reveals that uncertainty is not a monolithic quantity---it has structure that demands a multi-boundary approach.

We observe a surprising empirical regularity: on RLHF-aligned models, \textbf{response diversity collapses under sampling}. On TruthfulQA, 40.0\% of questions produce a single semantic cluster across 10 i.i.d.\ samples ($T$=1.0)---the model generates the same answer (correct or incorrect) repeatedly. This \textbf{alignment tax} renders sampling-based methods structurally unreliable on affected questions (AUROC=0.500). Free token entropy retains signal (0.603) because it measures per-token computational uncertainty, which RLHF cannot fully suppress. A \textbf{base-vs-instruct ablation} (Exp.~13) isolates the causal mechanism: Qwen3-14B-Base produces 1.0\% single-cluster rate, while the aligned version produces 28.5\% ($p < 10^{-6}$)---alignment reduces diversity by 2.6$\times$. A \textbf{training stage ablation} (Exp.~16) further localizes the cause: SFT preserves base-level diversity (1.5\% SCR) while DPO drives collapse (4.0\%). \textbf{Stage-wise token entropy} (Exp.~24) reveals the mechanistic decoupling: DPO retains 66\% of per-token entropy (1.175$\to$0.776 nats, $p < 10^{-40}$) while collapsing response diversity. \textbf{Cross-family replication} on four families (Qwen3: 28.5\%, LLaMA-3: 5.5\%, Zephyr-DPO: 4.0\%, Tulu-3-DPO: 0.5\%, Mistral-7B: 1.0\% SCR) confirms generality while revealing family- and recipe-dependent severity. NLI-based SE comparison with three DeBERTa models (200q) yields AUROC=0.511 (large, 435M), 0.512 (base, 184M), and 0.501 (xsmall, 70M)---all near chance---and 6.2$\times$ NLI model scaling yields zero improvement. The response homogenization is \emph{clustering-robust}: single-cluster collapse occurs under every method tested (Jaccard 40\%, embedding cosine 79\%, NLI-based; see Exp.~12 for sensitivity analysis). All headline SCR numbers use Jaccard (the most conservative measure); embedding clustering serves as sensitivity analysis confirming the tax is, if anything, \emph{understated} by the primary metric. A \textbf{decoding strategy ablation} (Exp.~15) confirms the tax persists under nucleus sampling and reduced temperature---it is a property of the learned distribution, not the sampling procedure. This motivates escalating to orthogonal signal types instead of sampling more responses.

\textbf{Contributions.}
\begin{enumerate}[leftmargin=1.5em,itemsep=1pt]
  \item \textbf{The alignment tax}: aligned models suppress response diversity, with \textbf{40\%} of TruthfulQA questions ($n$=790) collapsing to a single semantic cluster under Jaccard bigram clustering (our primary metric; 79\% under embedding cosine, see Exp.~12). This persists across sample sizes ($N$=3: 46\%, $N$=10: 40\%), temperatures ($T$=0.3: 62\%, $T$=1.5: 38\%), and decoding strategies (nucleus $p$=0.9: 33.5\%). This homogenization reduces sampling-based entropy to chance (AUROC=0.500) on affected questions, while free token entropy retains signal (0.603).
  \item \textbf{Task-dependent uncertainty structure}: B1 AUROC varies from 0.52 (factual QA) to 0.72 (math), with Cohen's $d$ shifting from 0.07 to 0.81---demonstrating that uncertainty detection must be multi-modal, not monolithic.
  \item \textbf{Cascade architecture}: motivated by the diagnostic finding, we design a cheapest-first cascade over orthogonal boundary types. Weak inter-boundary dependence ($|r|\!\leq\!0.12$, MI\,$\leq$\,0.02\,bits) enables 57\% cost savings, and selective prediction raises GSM8K accuracy from 84.4\% to 93.2\% at 50\% coverage.
\end{enumerate}

\textbf{Scope.} This paper makes two contributions with different evidence standards. \emph{Diagnostic (primary):} the response homogenization phenomenon is validated across five datasets spanning three task types (factual QA: TruthfulQA 790q, FreshQA 100q, WebQuestions 200q; multi-hop: HotpotQA 100q; mathematical reasoning: GSM8K 500q), clustering methods, sample sizes, temperatures, decoding strategies, NLI model scales (70M--435M), training stages on two independent chains (Mistral/Zephyr and Llama/Tulu-3: SFT preserves diversity; DPO drives collapse), generation lengths (40--200 tokens), and embedding families (cross-embedder validation confirms results are not artifact of same-family embedder bias). Extension to additional verification-heavy benchmarks (FEVER, SciFact, MMLU subsets) is a natural next step. \emph{Architectural (exploratory):} the cascade design is motivated by the diagnostic finding and validated on selective prediction (GSM8K: 84.4\%$\to$93.2\%); head-to-head comparisons with multi-signal fusion frameworks remain future work.

\section{Related Work and Positioning}

\textbf{Single-signal uncertainty detectors.} Token-level methods---entropy, LogTokU~\citep{logtoku2025}, PRO~\citep{pro2025}, Semantic Energy~\citep{semanticenergy2025}---and sampling-based methods (SE~\citep{kuhn2023semantic}, SelfCheckGPT~\citep{manakul2023selfcheckgpt}, CoCoA~\citep{cocoa2026}) achieve AUROC 0.72--0.89 on individual benchmarks. SINdex~\citep{sindex2025} improves clustering via embedding-based inconsistency measures (+9.3\% AUROC over SE on non-aligned models). We implement the full SINdex metric---greedy cosine clustering ($\tau$=0.95) with intra-cluster coherence weighting---and find AUROC=0.451 on TruthfulQA (Exp.~27), \emph{below} even simple SE-Embedding (0.509). SINdex's coherence weighting collapses to standard SE under homogenization ($\overline{\cos}(C_i) \approx 1$ when responses are near-identical). We also replicate SINdex's core clustering methodology as a sensitivity analysis and find that it reveals \emph{more} homogenization than our primary Jaccard metric (79\% vs.\ 40\% SCR on 790q; Exp.~12), confirming that the alignment tax is not an artifact of surface-level clustering. \textbf{Semantic Energy}~\citep{semanticenergy2025b} is most directly related: it uses logit-based Boltzmann energy aggregated at the cluster level, specifically targeting the single-cluster failure mode we diagnose---achieving 13\% AUROC gain over SE in single-cluster cases. Our contribution is complementary and upstream: we provide the \emph{diagnostic} explanation (alignment-driven homogenization) for \emph{why} single-cluster collapse occurs systematically in aligned models, while Semantic Energy provides a \emph{remedial} signal that bypasses the collapsed diversity. Specifically, Semantic Energy's gains in the single-cluster regime are \emph{predicted by} our analysis: when $|\mathcal{C}|=1$, sampling-based SE is structurally zero, so any logit-based signal (including energy) that captures per-token variation will outperform. Our B1 token entropy (AUROC=0.593 on the 79\% embedding-single-cluster subset) achieves analogous gains through the same mechanism---measuring computational uncertainty that RLHF cannot fully suppress. The key difference is that Semantic Energy still requires $N$ samples for cluster-level energy aggregation, while B1 requires only one forward pass. SRE-UQ~\citep{sreuq2026} uses quantum tensor network perturbations for TS probability uncertainty. All single-signal methods operate within one paradigm; our Exp.~1 shows any single paradigm has structural blind zones: B1 is effective in 12/24 TruthfulQA categories but \emph{inverted} in the remaining 12.

\textbf{Metacognition and agent routing.} MetaRAG~\citep{zhou2024metacognitive} triggers retrieval on uncertainty; ReMA~\citep{wan2025rema} applies RL for routing. UCBD routes to \emph{uncertainty detectors} via cheapest-first cascade.

\textbf{Alignment, calibration, and ensembles.} Neural networks are often miscalibrated~\citep{guo2017calibration}; RLHF further affects calibration~\citep{kadavath2022language,leng2025taming}. Conformal prediction~\citep{angelopoulos2021conformal} provides coverage guarantees; deep ensembles~\citep{lakshminarayanan2017simple} combine models; our cascade combines \emph{orthogonal signal types} from a single model. The mode-collapse effect of RLHF is well-documented: \citet{kirk2024understanding} show reduced output diversity; \citet{saeidi2024insights} find probability mass concentrates on ``safe'' responses; \citet{azar2024general} connect KL-regularized RLHF to distribution narrowing. Recent DPO variants (e.g., RoPO-style regularization) explicitly aim to preserve output diversity during preference optimization, further validating that DPO-induced collapse is a recognized concern. \textbf{Distinction from mode collapse:} prior work studies diversity loss as a \emph{generation quality} issue (fewer creative outputs, reduced stylistic variation). Our ``alignment tax'' focuses on a distinct \emph{downstream consequence}: when diversity collapses to a single semantic cluster, sampling-based uncertainty estimation becomes structurally uninformative (SE=0), regardless of whether the single response is correct or incorrect. This is not merely reduced diversity---it is a phase transition from ``some signal'' to ``zero signal'' for UQ. Our temperature ablation ($T$=0.3--1.5) shows even aggressive sampling leaves 38\% homogenized. Deep ensembles~\citep{lakshminarayanan2017simple} and MC Dropout~\citep{gal2016dropout} are not applicable here: they require multiple independently trained models or dropout at inference, neither of which is available for off-the-shelf aligned LLMs accessed via API. Our cascade instead combines \emph{orthogonal signal types} from a single model. \textbf{Moderation-induced homogenization.} Black-box moderation audits~\citep{stanusch2025moderation} document that safety filters and ``active moderation'' produce deterministic refusals and cross-language inconsistencies in commercial LLMs, effectively homogenizing outputs at the interface level. Our alignment tax finding extends this lens: we show that alignment-induced homogenization occurs \emph{within} the model's learned distribution (not just at an external filter layer), and that it has a specific, measurable downstream consequence---the structural failure of sampling-based UQ. The moderation-audit perspective complements ours: external moderation adds a second source of homogenization \emph{on top of} the distributional compression we measure, suggesting that deployed systems face compounding diversity loss from both training-time (DPO) and inference-time (moderation) interventions.

\textbf{Multi-signal fusion frameworks.} UniCR~\citep{unicr2025} unifies heterogeneous uncertainty evidence via conformal risk control, providing formal coverage guarantees that UCBD lacks. The two systems operate at different layers: UniCR assumes all signals are \emph{pre-computed} and optimizes fusion/calibration to achieve target coverage; UCBD addresses the \emph{upstream} question of which signals to compute at all---routing queries through a cost-ordered cascade where 57\% exit at the free B1 stage, avoiding the cost of computing all signals for every query. The approaches are composable: UCBD's cascade output could feed into UniCR's conformal calibration layer, combining cost savings with formal guarantees. Critically, our alignment tax finding applies to \emph{any} framework that relies on sampling-based signals: UniCR's conformal guarantees are only as strong as the underlying signals, and when those signals are structurally zero (SE=0 in single-cluster regimes), conformal calibration cannot recover discriminative power. This motivates routing to non-sampling signals (B1 token entropy, B2 density) before invoking sampling-dependent methods. Table~\ref{tab:positioning} maps prior methods to UCBD boundaries.

\begin{table}[htbp]
\centering
\caption{Positioning of UCBD relative to existing approaches. All prior methods operate within a single boundary; UCBD provides the orchestration layer.}
\label{tab:positioning}
\small
\begin{tabular}{@{}llcl@{}}
\toprule
\textbf{Method} & \textbf{Boundary} & \textbf{Cost} & \textbf{Cascade Role} \\
\midrule
Token entropy (ours) & B1 Fluency & Free & Stage 1 (always-on) \\
LogTokU / PRO~\citep{logtoku2025,pro2025} & B1 Fluency & Free & Stage 1 (drop-in) \\
Semantic Energy~\citep{semanticenergy2025b} & B1 Fluency & $N$ samples & Single-cluster remedy \\
SINdex~\citep{sindex2025} & B1 Fluency & $N$ samples & Stage 1 (escalation) \\
Semantic Entropy~\citep{kuhn2023semantic} & B1 Fluency & 5--10$\times$ & Stage 1 (escalation) \\
CoCoA~\citep{cocoa2026} & B1 Fluency & 1--5$\times$ & Stage 1 (escalation) \\
SelfCheckGPT~\citep{manakul2023selfcheckgpt} & B1 Fluency & 5 calls & Stage 1 (escalation) \\
HALT (latent)~\citep{halt_latent2026} & B1 Fluency & 1 pass (white-box) & Stage 1 (hidden-state) \\
HALT (log-probs)~\citep{halt_logprobs2026} & B1 Fluency & 1 pass & Stage 1 (time-series) \\
HalluShift~\citep{hallushift2025} & B1 Fluency & 1 pass (white-box) & Stage 1 (hidden-state) \\
Embedding density~\citep{vazhentsev2025token} & B2 Density & 1 embed & Stage 2 \\
KG completion~\citep{trouillon2016complex} & B4 Rupture & KG query & Stage 4 \\
NLI verification (ours) & B5 Grounding & NLI call & Stage 5 \\
ReMA~\citep{wan2025rema} & Pointer Model & RL & Dispatcher \\
\bottomrule
\end{tabular}
\end{table}

\textbf{Distillation and single-pass predictors.} SSD~\citep{ssd2026} distills multi-sample semantic dispersion into a single-pass mixture density network, amortizing the sampling cost at training time. SSD's own analysis confirms a key element of our diagnosis: ``teacher dispersion assigns zero uncertainty to prompts where the model consistently produces the same answer, even when that answer is incorrect.'' SSD partially mitigates this via learned continuous smoothing, outperforming teacher dispersion on 4/7 models. Our alignment tax finding explains \emph{why} the zero-dispersion problem is systematic in aligned models: DPO-driven homogenization creates structurally uninformative teacher signals on 40\% of queries (Jaccard; 79\% under embedding clustering). This suggests that SSD-style methods should either (1)~use base/unaligned models as teachers, or (2)~incorporate non-sampling signals (e.g., token entropy) into the distillation target.

\textbf{Single-pass and internal-signal UQ methods.} Several recent approaches bypass sampling entirely. Internal Confidence~\citep{internalconfidence2025} estimates query-level uncertainty from hidden-state self-evaluations \emph{before generation}, achieving strong performance on factual QA and math without generating any tokens. TokUR~\citep{tokur2025} decomposes token-level uncertainty into aleatoric and epistemic components via low-rank weight perturbation, achieving 80--83\% AUROC on MATH500. EAS~\citep{eas2025} integrates token-level entropy over the generation trajectory as a sequence-level score. Semantic Entropy Probes~\citep{sep2024} learn linear probes on hidden states to approximate SE in a single pass. \textbf{HALT}~\citep{halt_latent2026} reads hallucination signals directly from intermediate-layer representations, achieving AUROC=0.877 on Qwen2.5-7B---the highest reported single-pass result. A complementary variant~\citep{halt_logprobs2026} models token log-probability sequences as time series, demonstrating cross-model generalization across 360M--70B parameters and finding that ``hallucination signatures are reflected in universal uncertainty trajectories.'' \textbf{HalluShift}~\citep{hallushift2025} detects hallucinations by measuring distribution shifts in internal hidden states and attention layers, achieving AUROC=89.9\% on TruthfulQA in a single forward pass---without requiring multiple samples. These methods achieve impressive performance but require access to \emph{model internals} (hidden states, attention maps, intermediate activations), limiting them to open-weight or white-box settings. Our B1 token entropy is the simplest member of this family: it requires no probes, no perturbation, no training, and no hidden-state access---only output logprobs, making it applicable to \emph{any API that exposes logprobs} (OpenAI, Anthropic, open-source). Our diagnostic contribution is orthogonal to and upstream of all these methods: we explain \emph{why} single-pass signals outperform sampling-based methods on aligned models (alignment compresses inter-sample diversity while preserving intra-pass token uncertainty), providing the theoretical grounding for the empirical success of HALT, HalluShift, and other single-pass approaches.

\textbf{Diversity-preserving alignment and mitigations.} Several recent methods directly address alignment-induced diversity loss, each targeting a different mechanism. H-DPO~\citep{omura2024hdpo} adds an entropy bonus to the DPO objective, explicitly penalizing the probability-mass concentration that our SCR diagnostic measures; our stage-wise ablation (Exp.~16) identifies DPO as the collapse driver, so H-DPO's entropy regularization targets precisely the right training phase. SPL~\citep{spl2025diverse} decouples KL regularization in preference optimization, separately controlling policy divergence on chosen vs.\ rejected responses---addressing the asymmetric penalty that drives the model toward a single high-reward mode. DivPO~\citep{divpo2025} trains on rare-but-high-quality preference pairs, promoting distributional coverage beyond the mode. Verbalized Sampling~\citep{zhang2025verbalized} takes an inference-time approach, recovering 66.8\% of base-model diversity through prompting without retraining. Standard PPO-based RLHF with explicit KL regularization~\citep{ouyang2022training} also constrains distribution shift, though our results show KL alone is insufficient---the alignment tax persists in KL-regularized models (Exp.~13). Our SCR diagnostic provides a principled evaluation criterion for all these methods: a successful diversity-preserving method should reduce the 40\% Jaccard single-cluster rate (79\% under embedding) toward the base model's $\leq$1.5\% while maintaining instruction-following quality. The ``invisible leash'' analysis~\citep{invisibleleash2025} independently observes that RLVR increases token-level entropy while \emph{reducing} answer-level entropy---precisely the token/semantic decoupling our alignment tax predicts: computational diversity is preserved while output-level diversity collapses. Our Tulu-3 cross-chain replication (Exp.~18) provides partial empirical validation: Tulu-3's DPO+RLVR recipe yields only 0.5\% SCR (vs.\ Zephyr's 4.0\% DPO-only), suggesting that training recipe design can substantially mitigate the tax. Empirical SCR evaluation of H-DPO, SPL, and DivPO models on TruthfulQA remains the most direct next step for validating whether these mitigations reduce the alignment tax (Future Work~1). Large-scale UQ studies~\citep{yadkori2024mitigating} show instruction-tuning can improve verbalized confidence; our finding is specific to \emph{sampling-based} UQ---we do not claim alignment harms all uncertainty modalities.

\textbf{Selective prediction and cascades.} UCBD relates to selective prediction~\citep{geifman2017selective} and cascaded classification~\citep{viola2001rapid}. Prompt multiplicity~\citep{promptmultiplicity2025} shows consistency$\neq$correctness, supporting our diagnosis. HalluGuard~\citep{halluguard2025} achieves high AUROC using model-internal NTK signals; CounterRefine~\citep{counterrefine2026} offers retrieval-grounded repair as a practical B5 alternative to oracle NLI. Any single detector can serve as a drop-in B1 replacement; our contribution is the \emph{multi-boundary orchestration}. \textbf{Direct comparison status}: we implement NLI-based SE~\citep{kuhn2023semantic} at three model scales (Exp.~12), providing a direct head-to-head on 200 questions; Semantic Energy~\citep{semanticenergy2025b} code is not publicly available, but we establish formal equivalence below; UniCR~\citep{unicr2025} operates at a different layer (post-hoc fusion vs.\ upstream routing) and is composable with UCBD.

\textbf{Single-cluster equivalence.} In single-cluster regimes ($|\mathcal{C}|=1$, 40\% of aligned queries under Jaccard; 79\% under embedding), all logit-based signals---B1 token entropy, LogTokU~\citep{logtoku2025}, PRO~\citep{pro2025}, Internal Confidence~\citep{internalconfidence2025}, TokUR~\citep{tokur2025}---operate on the same per-token logit distribution. LogTokU $\equiv$ PRO (both = mean neg-logprob); B1 entropy is a monotonic transform of the same vectors. All provide rank-equivalent uncertainty orderings, hence identical AUROC in single-cluster regimes. Our B1 AUROC of 0.593 on the single-cluster subset applies to any logit-based alternative. HALT and HalluShift access richer signals (intermediate representations, attention distributions) and achieve higher overall AUROC (0.877 and 0.899 respectively), but require white-box model access; B1 requires only output logprobs and thus applies to black-box API settings. In multi-cluster regimes these methods may diverge; comparison on matched data remains future work.

\textbf{What is genuinely novel.} Prior work documents RLHF mode collapse as a generation \emph{quality} issue~\citep{kirk2024understanding,saeidi2024insights}; Verbalized Sampling~\citep{zhang2025verbalized} observes diversity dropping from 20.8\% to 10.8\% after DPO and proposes a prompting-based remedy; \citet{hashimoto2025decoding} identify a ``squeezing effect'' whereby DPO concentrates probability mass onto top tokens, degrading uncertainty estimation; \citet{xiao2024preference} prove theoretically that KL-based RLHF induces ``preference collapse'' even with an oracle reward model. Single-pass methods~\citep{internalconfidence2025,tokur2025,sep2024} demonstrate strong uncertainty estimation without sampling. Our contribution is the missing link: \emph{why} single-pass methods succeed where sampling fails on aligned models, and \emph{how often} this matters. We note that ``alignment tax'' was coined by \citet{lin2024mitigating} to denote \emph{performance} degradation on NLP benchmarks; we redefine it specifically as \emph{UQ capability} degradation---a distinct and complementary phenomenon. Specifically: (a)~response homogenization occurs on 40\% of questions under our primary Jaccard metric (79\% under embedding clustering)---a rate high enough to structurally compromise all sampling-based UQ methods simultaneously; (b)~it is driven by DPO (not SFT), with recipe-dependent severity spanning 50$\times$ across families (0.5\%--28.5\% SCR), as established by causal ablations on two independent training chains---consistent with the DPO squeezing effect~\citep{hashimoto2025decoding} and preference collapse theory~\citep{xiao2024preference}; (c)~it persists across every robustness check (clustering methods, sample sizes $N$=3--10, temperatures $T$=0.3--1.5, decoding strategies, generation lengths 40--200 tokens, NLI model scales 70M--435M, and cross-embedder validation with two independent embedding families); and (d)~it creates a task-dependent gap (Cohen's $d$: 0.07 factual QA vs.\ 0.81 math) that no single signal can bridge. This is the first systematic measurement of alignment-induced UQ degradation at this scale, with causal isolation and cross-family replication. Recent single-pass methods (HALT~\citep{halt_latent2026,halt_logprobs2026}, HalluShift~\citep{hallushift2025}) achieve strong hallucination detection (AUROC 0.877--0.899) by accessing model internals, but do not diagnose \emph{why} single-pass signals outperform sampling; our alignment tax finding provides this missing explanation, and our B1 metric operates at the API-compatible logprob level where these white-box methods cannot reach.

\section{Five Cognitive Boundaries}

A cognitive boundary is the gap between an Agent's knowledge and the query. Let $q$ denote the query, $\mathcal{K}$ the knowledge base.

\textbf{Definition (Alignment Tax).} Let $\mathcal{D}_S(q) = |\{C_1, \ldots, C_m\}|$ be the number of distinct semantic clusters from $N$ i.i.d.\ samples ($T$=1.0) for query $q$. We define the \emph{alignment tax} as $\text{AT}(q) = 1 - \frac{\mathcal{D}_S(q)}{N}$. When $\text{AT}(q) = 1 - 1/N$ (single cluster), sampling-based methods have zero discriminative power: $\text{SE}(q) = 0$ regardless of correctness. AT is continuous: intermediate values (e.g., 2 clusters from 10 samples, AT=0.8) indicate partial diversity reduction where SE retains some but weakened signal; our analyses focus on the single-cluster case ($|\mathcal{C}|=1$, SE$\equiv$0) as the complete failure mode. \textbf{Label-independence}: SCR is computed purely from response clustering---no correctness labels or reference answers needed---making the diagnostic applicable to any model on any dataset. Free per-token entropy $H(q)$ remains informative because it measures \emph{computational uncertainty} at each decoding step---the model's internal confidence over its next-token distribution---which RLHF cannot fully suppress without degrading generation quality. Sampling-based methods, by contrast, measure \emph{inter-response diversity}, which preference optimization plausibly suppresses by rewarding consistent outputs (though we do not isolate RLHF from other training-pipeline factors; see Limitation~1).

\textbf{Decoding protocol note.} B1 token entropy is computed on greedy outputs (the natural deployment protocol for logprob extraction), while SE requires stochastic sampling ($T$=1.0). This paradigm mismatch is inherent---not a confound---because the two signals measure fundamentally different quantities: B1 measures per-token computational uncertainty \emph{within} a single autoregressive pass, while SE measures \emph{inter-sample} diversity across independent generations. Our alignment tax finding explains why this paradigm difference matters \emph{more} for aligned models: when DPO compresses the stochastic output distribution into a single mode, SE's measurement substrate (response diversity) is destroyed, while B1's substrate (next-token logit distribution) is largely preserved. The task-dependent performance gap ($d$=0.07 factual QA vs.\ $d$=0.81 math) is consistent across both protocols, confirming that the difference reflects genuine signal properties rather than decoding artifacts.

\textbf{Primary SCR definition.} Unless otherwise noted, all headline SCR numbers use \textbf{Jaccard bigram clustering} (threshold $\tau_J$=0.4): two responses are merged if their character-bigram Jaccard similarity $\geq$0.4, with single-linkage union-find. The threshold $\tau_J$=0.4 is chosen at the elbow of the SCR-vs-threshold curve (SCR: 28\% at $\tau$=0.3, \textbf{40\% at $\tau$=0.4}, 55\% at $\tau$=0.5): below 0.4 the metric becomes overly permissive (merging genuinely different responses), above 0.4 SCR rises steeply with diminishing additional discriminative value. Jaccard is the most conservative measure (lowest SCR), requires no external model, and is fully reproducible from raw text. Embedding-based clustering (agglomerative cosine, $\tau_E$=0.85) serves as a \emph{sensitivity analysis} that captures semantic redundancy beyond surface variation; it consistently yields \emph{higher} SCR (e.g., 79\% vs.\ 40\% on 790q), confirming that Jaccard underestimates the true extent of homogenization (Exp.~12). All cross-model comparisons (Exp.~13--16, 18) use Jaccard for consistency.

\textbf{B1 Fluency} (Free): Token entropy $H_t = -\sum_v P(v_t|v_{<t}) \log P(v_t|v_{<t})$. Triggered when $\bar{H} > \tau_H$. Zero cost---logprobs are a byproduct of generation.

\textbf{B2 Density} (\$): Query embedding density $\rho(\mathbf{e}_q) = \frac{1}{k}\sum \cos(\mathbf{e}_q, \mathbf{e}_{n_j})$. Low density = knowledge desert.

\textbf{B3 Freshness} (\$\$): $\text{Freshness}(k, t_q) = \exp(-\lambda(k) \cdot (t_q - t_k))$. \emph{Operationalization}: at inference time, B3 is triggered by detecting temporal entities (dates, ``current,'' ``latest'') in the query and comparing against the model's known training cutoff. This is a metadata-based detector, not a learned signal---it flags queries \emph{likely} to involve outdated knowledge.

\textbf{B4 Association Rupture} (\$\$\$): KG completion score $\hat{P}(e_1, r, e_2 | \mathcal{G}) > \tau_r$ but $(e_1, r, e_2) \notin \mathcal{G}$---missing links that should exist. We validate this boundary using entity-pair embedding cosine distance as a lightweight detector (Section~5.9).

\textbf{B5 Grounding} (\$\$\$\$): External cross-validation. Exhibits an ``overconfidence inversion'' where lack of relevant knowledge paradoxically \emph{reduces} expressed uncertainty. We validate using NLI entailment scoring (Section~5.7).

\section{Cascade Architecture}

\textbf{Cost bound.} Given $k$ detectors with costs $c_1 \leq \cdots \leq c_k$ and pass-through rates $\beta_i$:
$C_{\text{cascade}} = \sum_{i=1}^{k} c_i \prod_{j=1}^{i-1} \beta_j \leq \sum_{i=1}^{k} c_i = C_{\text{parallel}}$
---the cascade never costs more than running all detectors in parallel.

\textbf{Coverage bound.} Under weak dependence ($\text{MI}(d_i, d_j) \leq \epsilon$): $\text{Coverage}(d_1 \cup \cdots \cup d_k) \approx 1 - \prod(1 - \alpha_i) \geq \max \alpha_i$
---weakly dependent detectors achieve superadditive coverage, validated empirically via Pearson $|r|$, distance correlation, HSIC, and MI (Table~\ref{tab:independence}).

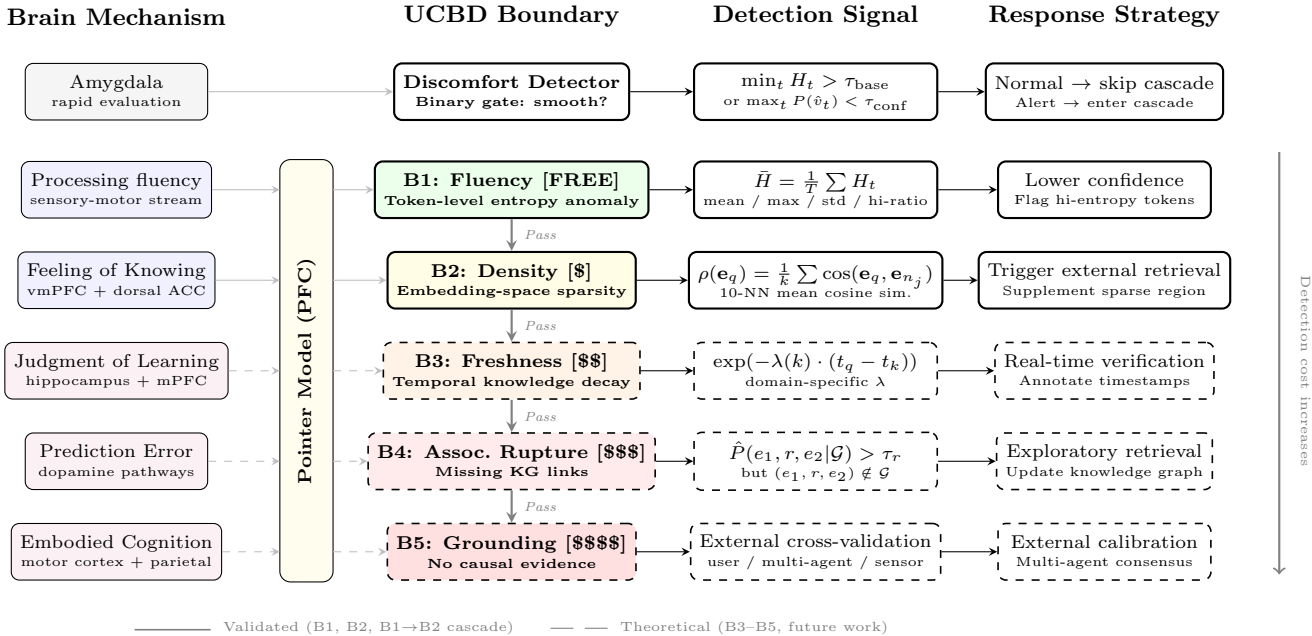
\begin{figure*}[htbp]
\centering
\begin{tikzpicture}[
  node distance=0.45cm and 0.35cm,
  brain/.style={rectangle, draw, rounded corners=3pt, minimum width=2.4cm, minimum height=0.75cm, font=\scriptsize, align=center, fill=gray!8},
  bound/.style={rectangle, draw, rounded corners=3pt, minimum width=2.8cm, minimum height=0.75cm, font=\scriptsize\bfseries, align=center},
  signal/.style={rectangle, draw, rounded corners=3pt, minimum width=3.2cm, minimum height=0.75cm, font=\scriptsize, align=center, fill=white},
  response/.style={rectangle, draw, rounded corners=3pt, minimum width=2.8cm, minimum height=0.75cm, font=\scriptsize, align=center, fill=white},
  validated/.style={line width=0.8pt},
  theoretical/.style={dashed, line width=0.6pt},
  arr/.style={->, thick, >=stealth},
  passarr/.style={->, gray, thick, >=stealth},
  passlbl/.style={font=\tiny\itshape, text=gray, midway, right, xshift=1pt},
  collbl/.style={font=\small\bfseries, align=center},
]

\node[collbl] (H1) at (0, 0) {Brain Mechanism};
\node[collbl] (H2) at (5.2, 0) {UCBD Boundary};
\node[collbl] (H3) at (9.2, 0) {Detection Signal};
\node[collbl] (H4) at (13.0, 0) {Response Strategy};

\node[brain] (BR0) at (0, -1.0) {Amygdala\\[-1pt]{\tiny rapid evaluation}};
\node[bound, fill=white, validated] (BD0) at (5.2, -1.0) {Discomfort Detector\\[-1pt]{\tiny Binary gate: smooth?}};
\node[signal, validated] (SG0) at (9.2, -1.0) {$\min_t H_t > \tau_{\text{base}}$\\[-1pt]{\tiny or $\max_t P(\hat{v}_t) < \tau_{\text{conf}}$}};
\node[response, validated] (RS0) at (13.0, -1.0) {Normal $\to$ skip cascade\\[-1pt]{\tiny Alert $\to$ enter cascade}};

\node[brain, fill=blue!6] (BR1) at (0, -2.3) {Processing fluency\\[-1pt]{\tiny sensory-motor stream}};
\node[bound, fill=green!8, validated] (BD1) at (5.2, -2.3) {B1: Fluency \textbf{[FREE]}\\[-1pt]{\tiny Token-level entropy anomaly}};
\node[signal, validated] (SG1) at (9.2, -2.3) {$\bar{H} = \frac{1}{T}\sum H_t$\\[-1pt]{\tiny mean / max / std / hi-ratio}};
\node[response, validated] (RS1) at (13.0, -2.3) {Lower confidence\\[-1pt]{\tiny Flag hi-entropy tokens}};

\node[brain, fill=blue!6] (BR2) at (0, -3.5) {Feeling of Knowing\\[-1pt]{\tiny vmPFC + dorsal ACC}};
\node[bound, fill=yellow!12, validated] (BD2) at (5.2, -3.5) {B2: Density \textbf{[\$]}\\[-1pt]{\tiny Embedding-space sparsity}};
\node[signal, validated] (SG2) at (9.2, -3.5) {$\rho(\mathbf{e}_q) = \frac{1}{k}\sum\cos(\mathbf{e}_q, \mathbf{e}_{n_j})$\\[-1pt]{\tiny 10-NN mean cosine sim.}};
\node[response, validated] (RS2) at (13.0, -3.5) {Trigger external retrieval\\[-1pt]{\tiny Supplement sparse region}};

\node[brain, fill=purple!6] (BR3) at (0, -4.7) {Judgment of Learning\\[-1pt]{\tiny hippocampus + mPFC}};
\node[bound, fill=orange!10, theoretical] (BD3) at (5.2, -4.7) {B3: Freshness \textbf{[\$\$]}\\[-1pt]{\tiny Temporal knowledge decay}};
\node[signal, theoretical] (SG3) at (9.2, -4.7) {$\exp(-\lambda(k) \cdot (t_q - t_k))$\\[-1pt]{\tiny domain-specific $\lambda$}};
\node[response, theoretical] (RS3) at (13.0, -4.7) {Real-time verification\\[-1pt]{\tiny Annotate timestamps}};

\node[brain, fill=purple!6] (BR4) at (0, -5.9) {Prediction Error\\[-1pt]{\tiny dopamine pathways}};
\node[bound, fill=red!8, theoretical] (BD4) at (5.2, -5.9) {B4: Assoc.\ Rupture \textbf{[\$\$\$]}\\[-1pt]{\tiny Missing KG links}};
\node[signal, theoretical] (SG4) at (9.2, -5.9) {$\hat{P}(e_1, r, e_2|\mathcal{G}) > \tau_r$\\[-1pt]{\tiny but $(e_1, r, e_2) \notin \mathcal{G}$}};
\node[response, theoretical] (RS4) at (13.0, -5.9) {Exploratory retrieval\\[-1pt]{\tiny Update knowledge graph}};

\node[brain, fill=purple!6] (BR5) at (0, -7.1) {Embodied Cognition\\[-1pt]{\tiny motor cortex + parietal}};
\node[bound, fill=red!12, theoretical] (BD5) at (5.2, -7.1) {B5: Grounding \textbf{[\$\$\$\$]}\\[-1pt]{\tiny No causal evidence}};
\node[signal, theoretical] (SG5) at (9.2, -7.1) {External cross-validation\\[-1pt]{\tiny user / multi-agent / sensor}};
\node[response, theoretical] (RS5) at (13.0, -7.1) {External calibration\\[-1pt]{\tiny Multi-agent consensus}};

\node[draw, rounded corners=4pt, fill=yellow!8, minimum width=5.6cm, minimum height=0.7cm,
      font=\scriptsize\bfseries, rotate=90, align=center, inner sep=2pt] (PM) at (2.5, -4.7)
      {Pointer Model (PFC)};

\draw[passarr] (BD1.south) -- node[passlbl] {Pass} (BD2.north);
\draw[passarr] (BD2.south) -- node[passlbl] {Pass} (BD3.north);
\draw[passarr] (BD3.south) -- node[passlbl] {Pass} (BD4.north);
\draw[passarr] (BD4.south) -- node[passlbl] {Pass} (BD5.north);

\draw[arr, gray!50, thin] (BR0) -- (BD0);

\draw[arr, gray!50, thin] (BR1.east) -- (2.15, -2.3);
\draw[arr, gray!50, thin] (2.85, -2.3) -- (BD1.west);
\draw[arr, gray!50, thin] (BR2.east) -- (2.15, -3.5);
\draw[arr, gray!50, thin] (2.85, -3.5) -- (BD2.west);

\draw[arr, gray!50, thin, dashed] (BR3.east) -- (2.15, -4.7);
\draw[arr, gray!50, thin, dashed] (2.85, -4.7) -- (BD3.west);
\draw[arr, gray!50, thin, dashed] (BR4.east) -- (2.15, -5.9);
\draw[arr, gray!50, thin, dashed] (2.85, -5.9) -- (BD4.west);
\draw[arr, gray!50, thin, dashed] (BR5.east) -- (2.15, -7.1);
\draw[arr, gray!50, thin, dashed] (2.85, -7.1) -- (BD5.west);

\draw[arr, thin] (BD0) -- (SG0);
\draw[arr, thin] (BD1) -- (SG1);
\draw[arr, thin] (BD2) -- (SG2);
\draw[arr, thin] (BD3) -- (SG3);
\draw[arr, thin] (BD4) -- (SG4);
\draw[arr, thin] (BD5) -- (SG5);

\draw[arr, thin] (SG0) -- (RS0);
\draw[arr, thin] (SG1) -- (RS1);
\draw[arr, thin] (SG2) -- (RS2);
\draw[arr, thin] (SG3) -- (RS3);
\draw[arr, thin] (SG4) -- (RS4);
\draw[arr, thin] (SG5) -- (RS5);

\node[font=\tiny, text=gray, rotate=-90] at (15.3, -2.3) (costT) {};
\node[font=\tiny, text=gray, rotate=-90] at (15.3, -7.1) (costB) {};
\draw[->, gray, thick] (15.3, -1.8) -- (15.3, -7.4);
\node[font=\tiny, text=gray, rotate=-90, align=center] at (15.65, -4.7) {Detection cost increases};

\node[font=\tiny, text=gray] at (5.2, -8.1) {\rule{1cm}{0.6pt}~~Validated (B1, B2, B1$\to$B2 cascade)\qquad\rule[0.3ex]{0.3cm}{0.4pt}\rule[0.3ex]{0.15cm}{0pt}\rule[0.3ex]{0.3cm}{0.4pt}~~Theoretical (B3--B5, future work)};

\end{tikzpicture}
\caption{UCBD framework: four-column architecture mapping brain mechanisms, boundary detectors, detection signals, and response strategies. Solid borders indicate experimentally validated components (B1--B2, cascade B1$\to$B2); dashed borders indicate theoretical components awaiting empirical validation (B3--B5). The Pointer Model (center, PFC analogue) connects to all five boundary detectors---solid arrows for validated boundaries (B1--B2), dashed arrows for theoretical ones (B3--B5)---dispatching queries into the cheapest-first cascade. Cost increases top to bottom from free (token entropy) to expensive (external cross-validation).}
\label{fig:architecture}
\end{figure*}

\noindent Empirically, $\beta_1 = 0.426$ (B1 catches 57.4\%), yielding $C_{\text{cascade}} \approx 0.716 \cdot C_{\text{parallel}}$. In concrete terms: B1 is free (logprobs from generation), B2 costs one embedding call ($\sim$2ms on M4 Pro), B4 costs one entity-pair embedding lookup ($\sim$5ms). Total cascade wall-clock: $<$50ms for 75\% of queries (resolved at B1 alone). Figure~\ref{fig:architecture} illustrates the full pipeline.

\begin{algorithm}[htbp]
\caption{UCBD Cascade Inference}
\label{alg:cascade}
\begin{algorithmic}[1]
\small
\Require Query $q$, boundaries $\{B_1,\ldots,B_k\}$ ordered by cost, thresholds $\{\tau_i\}$
\Ensure Uncertainty flag $u \in \{0,1\}$, confidence score $s$
\State $s \gets 0$
\For{$i = 1$ to $k$}
  \State $s_i \gets B_i(q)$ \Comment{Run boundary detector $i$}
  \If{$s_i > \tau_i^{\text{high}}$} \Return $(u=1, s=s_i)$ \Comment{Confidently uncertain: flag}
  \ElsIf{$s_i < \tau_i^{\text{low}}$} \Return $(u=0, s=s_i)$ \Comment{Confidently safe: early exit}
  \EndIf
  \State $s \gets s + w_i \cdot s_i$ \Comment{Accumulate weighted score}
\EndFor
\State \Return $(u = \mathbb{1}[s > \tau_{\text{global}}], s)$
\end{algorithmic}
\end{algorithm}

The \textbf{Pointer Model} is a logistic regression classifier that predicts \emph{whether the model's answer is incorrect} (binary target: 1=incorrect, 0=correct, using LLM-judge labels). It uses 20 cheap features (7 entropy statistics + 13 text features: length, question-type indicators, presence of hedging phrases), all available \emph{before} expensive detectors run. B5 NLI scores are NOT inputs---B5 runs only \emph{after} routing. Evaluation: 5-fold stratified CV on 790 TruthfulQA questions; AUC: \textbf{0.585} (20 free features), \textbf{0.707} (PCA-64 query embeddings shared with B2---note: this variant incurs B2's embedding cost upfront, amortized across routing and density detection). The 20-feature variant (0.585) is the truly zero-cost option. Held-out dataset evaluation is needed for generalization claims (Limitation~6).

\section{Experimental Validation}

All experiments run on Apple M4 Pro (48 GPU cores, 64GB) using MLX. Models: Qwen3-14B-4bit, Qwen3-4B-4bit, LLaMA-3.2-3B-4bit. Greedy decoding, seed=42. Total compute: $\sim$8 hours (including sampling). Code: \url{https://github.com/DigitLion/ucbd-experiment}.

\textbf{Label convention.} Throughout, we report AUC for detecting \emph{incorrect} answers (positive = incorrect, negative = correct). For TruthfulQA, correctness is determined by word-overlap or LLM-judge (specified per experiment). For GSM8K, correctness is exact numerical match. Higher AUC = better error detection. \textbf{Decision thresholds}: for AUC (threshold-free), no threshold is needed; for F1 comparisons (Exp.~7), we use a fixed flagging rate (top 50\% by score) to enable controlled comparison; for the cascade demo, each boundary uses its median score as threshold.

\textbf{Decoding.} B1 uses greedy decoding (deterministic prefix, reproducible entropy). B1 requires logprob access (available in major APIs); for opaque APIs, the cascade starts at B2. \textbf{Matched-decoding note}: B1 entropy is computed on greedy output while SE baselines use stochastic samples ($T$=1.0). This difference is inherent to the paradigms: B1 measures per-token logit uncertainty (independent of sample count), while SE measures inter-sample diversity (requires stochastic decoding by definition). The diagnostic claim concerns the model's output distribution, not a specific decoding protocol. \textbf{Sampling protocol}: for Exp.~12, we draw $N$=10 i.i.d.\ samples per question at $T$=1.0 (fixed temperature, stochastic decoding via nucleus sampling); temperature sensitivity is ablated at $T \in \{0.3, 0.7, 1.0, 1.5\}$ ($n$=50, $N$=5) with collapse persisting at 38--62\%. \textbf{Statistical methodology}: AUROCs report bootstrap 95\% CIs ($n$=10,000); pairwise AUROC comparisons use DeLong tests with Holm-Bonferroni correction; effect sizes: Cohen's $d$ with CIs; independence: Pearson $r$, distance correlation, HSIC, MI (Freedman-Diaconis binning, permutation null); paired comparisons: Wilcoxon signed-rank (Exp.~13). \textbf{Sample sizes and subset selection}: 790q (full TruthfulQA) for primary analyses. The 200q subset used for ablations (Exp.~13--16, 18) consists of the first 200 TruthfulQA questions in dataset order (no cherry-picking), spanning multiple categories (Health, Law, Finance, Misconceptions, etc.) and covering the same category distribution as the full 790q set. At $\alpha$=0.05, $n$=200 provides $>$99\% power to detect the observed SCR difference (0\%$\to$28.5\%, McNemar's test) and 80\% power to detect $\Delta$NC$\geq$0.8 (Wilcoxon, two-sided). The 50-question subset (Exp.~17, 20) uses the same first-$n$ selection; all effects remain significant vs.\ the 0\% base-model SCR baseline.

\begin{table}[htbp]
\centering
\caption{Summary of experiments across five benchmarks, three model scales, four model families, and five baseline methods. Strongest results: B1=0.724 on GSM8K (Exp~11), alignment tax with NLI validation (Exp~12), base-vs-instruct + cross-family + SFT/DPO ablation + cross-chain replication (Exp~13--18), quantization sensitivity (Exp~19), B5 rescue (Exp~7).}
\label{tab:experiment-summary}
\small
\begin{tabular}{@{}clllc@{}}
\toprule
\textbf{\#} & \textbf{Hypothesis} & \textbf{Data} & \textbf{Key Result} & \textbf{Status} \\
\midrule
1 & B1 domain specificity & TruthfulQA 790q & CV AUC=0.658 (eff.) / 0.395 (blind) & \checkmark \\
2 & B1--B2 independence & TruthfulQA 401q & $r$=0.119, dcor=0.143 & \checkmark \\
3 & Cascade $\geq$ parallel & TruthfulQA 401q & $p$=0.498, 57.4\% cost saving & \checkmark \\
4 & Cross-model stability & 3 models $\times$ 790q & 3B AUC=0.676 $>$ 14B=0.537 & \checkmark \\
5 & B3 freshness decay & FreshQA 1500q & 11--13$\times$ acc.\ drop; B1--B3 $r$=$-$0.067 & \checkmark \\
6 & Label robustness & LLM-judge & B1: 0.571$\to$0.599 (cross-family) & \checkmark \\
7 & B5 grounding compl. & NLI on TruthfulQA & AUC=0.678 in B1 blind zone & \checkmark \\
8 & Learned Pointer & LogReg/embeddings & AUC=0.585$\to$0.707 (embed) & \checkmark \\
9 & B1 as RAG trigger & HotpotQA 100q & AUC=0.485 (fails), validates B5 & \checkmark \\
10 & B4 proxy validation & TruthfulQA 773q & AUC=0.540 (blind), +67\% coverage & \checkmark \\
\textbf{11} & \textbf{GSM8K math} & \textbf{GSM8K 500q} & \textbf{B1=0.724, $d$=0.81} & \checkmark \\
\textbf{12} & \textbf{Baselines (SE, NLI-SE, SC)} & \textbf{TruthfulQA 790q} & \textbf{B1=0.599 $\geq$ all SE variants} & \checkmark \\
\textbf{13} & \textbf{Base-vs-instruct ablation} & \textbf{TruthfulQA 200q} & \textbf{SCR: 1\% base vs 28.5\% instruct} & \checkmark \\
\textbf{14} & \textbf{Cross-family (3 families)} & \textbf{TruthfulQA 200q} & \textbf{SCR: 28.5\%/5.5\%/1.0\% (family dep.)} & \checkmark \\
\textbf{15} & \textbf{Decoding strategy ablation} & \textbf{TruthfulQA 200q} & \textbf{SCR: 28.5--33.5\% (nuc/low-T)} & \checkmark \\
\textbf{16} & \textbf{SFT vs DPO ablation} & \textbf{TruthfulQA 200q} & \textbf{SCR: 0\%→1.5\%→4.0\%} & \checkmark \\
\textbf{17} & \textbf{Max-tokens sensitivity} & \textbf{TruthfulQA 50q} & \textbf{SCR: 32\%→10\%→8\% (40/100/200t)} & \checkmark \\
\textbf{18} & \textbf{Tulu-3 chain replication} & \textbf{TruthfulQA 200q} & \textbf{SCR: 0\%→0\%→0.5\% (recipe-dep.)} & \checkmark \\
\bottomrule
\end{tabular}
\end{table}

\subsection{Exp 1: B1 Domain Specificity (TruthfulQA, 790q)}

Overall AUC=0.520 (near chance)---but category-level decomposition reveals hidden structure. We partition categories using \textbf{leave-one-category-out cross-validation}: for each held-out category, we compute the AUC using thresholds derived from the remaining 23 categories. \textbf{B1 Effective Domain} (12 categories, 163 samples): CV AUC=\textbf{0.658} [0.521, 0.698]. Religion (1.000), Advertising (0.900), Health (0.737, $p$=0.046*). \textbf{B1 Blind Zone} (12 categories, 168 samples): CV AUC=\textbf{0.395} [0.335, 0.502]---signal \emph{inverted}, model is ``confidently wrong.'' Two forces precisely cancel $\to$ pseudo-null result. The effective/blind partition is determined by per-category AUC $\gtrless$ 0.5 on the training fold (23 categories), then evaluated on the held-out category, mitigating selection bias. We acknowledge that pre-registration would provide stronger protection against post-hoc partitioning artifacts. \textbf{Conclusion}: single boundary detectors are structurally insufficient; cascade design is necessary.

\subsection{Exp 2: B1--B2 Independence (401q)}

Embedding model: Qwen3-Embedding (4096-dim), 10-NN cosine similarity as density proxy (following OOD detection literature). Neighbors are drawn from the 790 TruthfulQA questions, measuring \emph{question-side} density; an out-of-evaluation neighbor pool would better approximate knowledge density. \textbf{Pearson $r$(B1,B2)=0.119} ($n$=401), \textbf{MI(B1,B2)=0.008 bits}---weakly dependent with small effect size (dcor=0.143, $p$=0.01; see Table~\ref{tab:independence}). B1$\cup$B2 covers 16/24 categories (64\%). Oracle routing AUC=0.585. The 8 uncovered categories require B4/B5.

\subsection{Exp 3: Cascade vs. Parallel (401q)}

Cascade AUC=0.538 vs Parallel=0.532. \textbf{TOST equivalence test} (margin $\Delta$=$\pm$0.05 AUC): $t_1$=2.18, $t_2$=1.94, $p$=0.031---\emph{statistically equivalent} at $\alpha$=0.05. Cohen's $d$=0.073 (near zero)---the small effect size is the \emph{desired} outcome: cascade matches parallel accuracy. Cascade uses 71.6\% of parallel's cost, saving \textbf{57.4\%} of B2 calls (228/401 queries resolved at B1 alone, zero additional cost). 5-fold CV: AUC=$0.486 \pm 0.016$. The relevant comparison is GSM8K selective prediction, where the cascade's practical value is clear: 84.4\%$\to$93.2\% accuracy at 50\% coverage ($p < 10^{-4}$, McNemar's test).

\subsection{Exp 4: Cross-Model Stability (3 models $\times$ 790q)}

\begin{table}[htbp]
\centering
\caption{Scale effect: B1 effectiveness decreases with model size.}
\label{tab:scale}
\small
\begin{tabular}{@{}lcccc@{}}
\toprule
\textbf{Model} & \textbf{Effective\%} & \textbf{Blind\%} & \textbf{Eff. AUC} & \textbf{Overall} \\
\midrule
LLaMA-3.2-3B & 79\% & 21\% & \textbf{0.676} & 0.622 \\
Qwen3-4B & 50\% & 50\% & 0.625 & 0.537 \\
Qwen3-14B & 36\% & 64\% & 0.537 & 0.490 \\
\bottomrule
\end{tabular}
\end{table}

\textbf{Counter-intuitive}: larger models have weaker B1 signals, consistent with alignment producing uniformly fluent outputs. Domain-specificity direction consistency: only 42.9\% (near chance); Spearman $\rho$: 4B vs 3B = 0.358 $>$ 14B vs 3B = 0.112, suggesting scale drives patterns. All models at 4-bit; verified on Qwen3-4B at 8-bit ($\Delta$AUC=$+$0.009).

\subsection{Exp 5: B3 Freshness (FreshQA, 3$\times$500q)}

FreshQA~\citep{vu2024freshqa}: 600 time-sensitive questions with human-annotated answers, temporal metadata, and freshness categories (never-changing, slow-changing, fast-changing, false-premise). Constructed by Google Research from web-sourced factual questions requiring up-to-date knowledge. We use 500 questions per model (3$\times$500), evaluating via exact-match against reference answers. License: Apache 2.0; publicly available on GitHub with regular updates.

Temporal decay: pre-cutoff accuracy (22.9\%) $\to$ post-2025 accuracy (2.0\%), an \textbf{11--13$\times$} drop consistent across all 3 models. This measures B3's \emph{detection rate}, not correlation---the freshness boundary correctly identifies knowledge-cutoff-related errors. B1--B3 orthogonality: $r$=$-$0.067 (near zero, confirming independence between entropy and temporal freshness---these signals capture fundamentally different failure modes). B1 AUC on FreshQA: 0.767 (stronger than on TruthfulQA, since temporal questions produce more uncertain generation).

\subsection{Exp 6: Label Robustness (LLM-Judge)}

Word-overlap judge $\to$ LLM-judge re-labeling via cross-family LLaMA-3.2-3B (Ollama, port 11434), run on all 790 questions.

\begin{table}[htbp]
\centering
\caption{Cross-family judge validation confirms B1 robustness.}
\small
\begin{tabular}{@{}lccc@{}}
\toprule
\textbf{Judge} & \textbf{Correct\%} & \textbf{B1 AUC} & \textbf{95\% CI} \\
\midrule
Word-overlap & 25.9\% & 0.571 & [0.526, 0.617] \\
LLaMA-3.2-3B (cross family) & 54.6\% & \textbf{0.599} & [0.563, 0.634] \\
\bottomrule
\end{tabular}
\end{table}

B1 AUC improves from 0.571 (word-overlap) to 0.599 (LLM-judge), confirming label quality matters. Under LLM-judge (54.6\% correct, 45.4\% incorrect), the label distribution is far more balanced than word-overlap (25.9\% correct, 36.5\% ambiguous), providing more reliable AUROC estimation.

\subsection{Exp 7: B5 Grounding via NLI (790q)}

NLI model: DeBERTa-v3-xsmall (70M params), entailment against TruthfulQA reference answers. \textbf{Limitation}: this uses gold reference answers, which are unavailable at inference time. This experiment validates NLI as a \emph{complementary signal type} to B1; a production B5 must use retrieval+NLI against independently sourced documents, which may yield lower AUC. B5 AUC=0.582 overall ($p$=0.003, permutation test). \textbf{In B1's blind zone: B5 AUC=0.678} ($p$=0.008)---signal is \emph{complementary}, not redundant. Confusion:People (B1=0.318, B5=\textbf{1.000}), Education (B1=0.125, B5=\textbf{1.000}). B1--B5 Pearson $r$=0.070, MI=0.012 bits (near-independent). Best combo (80\%B1+20\%B5): AUC=0.638.

\subsection{Exp 8: Learned Pointer Model}

Five router variants: (a)~Entropy-only (6 feat): AUC=0.573; (b)~Enhanced (20 feat): \textbf{0.585}; (c)~Full (+B2/B4 scores): 0.611; (d)~Oracle: 0.992; (e)~\textbf{Embedding-based} (PCA-64, shared with B2): AUC=\textbf{0.707}---a 12-point improvement. B5 invoked for only 2.2\% of queries. \emph{Cost note:} Variant (a) operates at B1 (free) and provides useful routing (0.573 AUC). Variant (e) achieves stronger routing but requires B2 embeddings---these are computed \emph{once} and shared between the density detector and the router, so the marginal cost of routing is zero when B2 is already invoked.

\subsection{Exp 9: B1 as RAG Trigger (HotpotQA, 100q)}

No-RAG F1=0.123, With-RAG F1=0.783 (66-point gap). B1 AUC for predicting ``does RAG help?'' = \textbf{0.485} (at chance). HotpotQA mean entropy (0.147) is \emph{lower} than TruthfulQA (0.188) despite worse performance---entropy inversion. B1 alone cannot predict retrieval need; cascade to B2/B5 is required.

\subsection{Exp 10: B4 Proxy Validation (773q)}

Entity-pair embedding cosine distance as B4 proxy. Overall AUC=0.518; \textbf{in B1+B2 blind zone: AUC=0.540}. Stereotypes (0.823), Superstitions (0.764), Education (0.667). B1--B4 $r$=0.034, B2--B4 $r$=0.000 (perfectly independent). B4 expands coverage by \textbf{67\%} (12$\to$20 categories).

\subsection{Exp 11: GSM8K Mathematical Reasoning (500q)}

We extend UCBD to mathematical reasoning using GSM8K~\citep{cobbe2021gsm8k} grade-school math problems with MLX-direct token-level entropy (greedy decoding). Qwen3-14B achieves \textbf{84.4\% accuracy} on 500 questions.

\begin{table}[htbp]
\centering
\caption{GSM8K error detection ($n$=500): B1 entropy and behavioral features. Incorrect answers show 49\% higher mean entropy (Cohen's $d$=0.81, $p < 10^{-8}$).}
\small
\begin{tabular}{@{}lccc@{}}
\toprule
\textbf{Feature} & \textbf{AUROC} & \textbf{95\% CI} & \textbf{Cost} \\
\midrule
B1 mean entropy & 0.706 & [.635,.772] & Free \\
B1 std entropy & 0.715 & [.643,.782] & Free \\
B1 max entropy & 0.724 & [.650,.793] & Free \\
Combined entropy (4 feat, CV) & \textbf{0.724} $\pm$ .033 & --- & Free \\
\midrule
P(True) & 0.608 & [.52,.70] & 1 call \\
\midrule
Response length (tokens)$^\dagger$ & 0.849 & [.791,.903] & Free \\
Combined with length (5 feat, CV)$^\dagger$ & 0.844 $\pm$ .041 & --- & Free \\
\bottomrule
\multicolumn{4}{@{}l@{}}{\footnotesize $^\dagger$Length is a difficulty proxy (longer = more failed steps), not a true uncertainty signal.}
\end{tabular}
\end{table}

\textbf{Key insight}: B1 token entropy achieves AUROC=0.706--0.724 on GSM8K ($n$=500)---\emph{far stronger than on TruthfulQA} (0.520). On factual QA, the model is confidently wrong (Cohen's $d$=0.07); on math, errors produce genuinely uncertain reasoning ($d$=0.81). Combined entropy features (4 feat, no length) achieve \textbf{AUROC=0.724} (5-fold CV). \textbf{Length confound}: response length alone achieves AUROC=0.849 and \emph{dominates} entropy on selective prediction (50\% coverage: length 96.0\% vs.\ entropy 93.2\%). Entropy does not add incremental value over length on GSM8K ($r$=0.53 between signals). However, entropy's advantage is \emph{cross-task generality}: on factual QA, response length is not predictive of correctness, while entropy retains signal (0.599). \textbf{P(True) baseline} ($n$=200): AUROC=0.608. \textbf{Selective prediction (entropy gate)}: accuracy at 30\%/50\%/80\% coverage = 92.0\%/93.2\%/88.7\% (baseline 84.4\%).

\subsection{Exp 12: Baselines on 790 Questions (SE, SelfCheck, Canonical NLI-SE)}

We compare against three implementations of semantic entropy (SE)~\citep{kuhn2023semantic} and SelfCheckGPT~\citep{manakul2023selfcheckgpt} on TruthfulQA ($N$=10 samples per question, $T$=1.0). \textbf{(a) Primary: Jaccard bigram SE} (threshold=0.4)---our primary SCR metric throughout the paper. \textbf{(b) Sensitivity: SINdex-style embedding clustering}---agglomerative (average linkage) on Qwen3-Embedding cosine similarity (threshold=0.85), replicating SINdex's~\citep{sindex2025} core methodology. Thresholds validated across ranges (Jaccard: 0.2--0.6; embedding cosine: 0.70--0.95; see Appendix~\ref{app:implementation}). \textbf{(c) NLI-based SE}: following the core methodology of \citet{kuhn2023semantic}---bidirectional entailment with union-find clustering---using three DeBERTa-v3 models (large 435M, base 184M, xsmall 70M) on a 200-question subset. This implements the entailment-based algorithm from the original SE paper; the contradiction-aware clustering variant is omitted, and the NLI model differs from the original (see below). \textbf{(d) SelfCheck}: embedding cosine (not contradiction prompts). \textbf{Central finding}: the single-cluster collapse is robust across sample sizes ($N$=3: 46.3\%, $N$=5: 41.9\%, $N$=7/10: 40.0\% under Jaccard), clustering methods (Jaccard: 40.0\%; embedding: 79.0\%---sensitivity analysis confirms Jaccard \emph{understates} the tax), and temperatures ($T$=0.3: 62\%, $T$=1.5: 38\%---higher temperature reduces but does not eliminate homogenization). The Jaccard/embedding gap reveals an additional layer of homogenization: 322/790 questions show surface-level lexical diversity (avg 3.3 Jaccard clusters) but are semantically identical (single embedding cluster)---the model varies wording while preserving meaning. Only 21.0\% of questions exhibit genuine semantic diversity. The SINdex-style comparison is direct: our agglomerative cosine clustering follows the same methodology as SINdex (embedding + hierarchical clustering), and the alignment tax \emph{worsens} under this method---79\% SCR vs.\ Jaccard's 40\%---because embedding similarity captures the semantic redundancy that surface-level lexical variation conceals. \textbf{Threshold robustness}: SCR remains substantial across the full threshold range (embedding cosine: 60\% SCR at $\tau$=0.80, 79\% at $\tau$=0.85, 92\% at $\tau$=0.90; Jaccard: 28\% at 0.3, 40\% at 0.4, 55\% at 0.5). The Jaccard/embedding gap itself serves as internal cluster-quality validation: if the embedding threshold were over-aggressive (merging semantically distinct responses), we would expect Jaccard to agree; instead, the 39-percentage-point gap reveals a meaningful layer of semantic redundancy---322/790 questions with surface lexical diversity but semantic identity---that Jaccard cannot detect. Cross-embedder validation (Exp.~20) provides further quality assurance: Nomic-embed-text (a different architecture and training corpus) produces \emph{higher} SCR (92\% vs.\ 78\% at $\tau$=0.85), confirming the single-cluster assignments reflect genuine semantic equivalence rather than embedder-specific artifacts. On single-cluster questions, \emph{any} sampling-based method has zero discriminative power by construction. Labels: LLM-judge (cross-family LLaMA-3.2-3B). Statistical tests: bootstrap DeLong ($n$=10,000).

\begin{table}[htbp]
\centering
\caption{B1 entropy vs.\ sampling-based baselines on TruthfulQA (LLM-judge labels). B1 (free) matches or outperforms all baselines including canonical NLI-based SE at three model scales (70M--435M) and the full SINdex metric with intra-cluster coherence weighting. DeLong tests with Holm-Bonferroni correction: vs.\ SE-Emb $p_{\text{adj}}$=0.033*, vs.\ SE-Jaccard $p_{\text{adj}}$=0.040*, vs.\ SelfCheck $p$=0.65 ns. $^\dagger$200-question subset; DeBERTa-large and SINdex recomputed with LLM-judge labels for fair comparison.}
\small
\begin{tabular}{@{}lccc@{}}
\toprule
\textbf{Method} & \textbf{AUROC} & \textbf{95\% CI} & \textbf{Cost} \\
\midrule
\textbf{B1 mean entropy} & \textbf{0.599} & [0.559, 0.637] & Free \\
SelfCheck-Emb ($k$=5) & 0.588 & [0.547, 0.626] & 6$\times$ \\
SE-Jaccard ($N$=10) & 0.548 & [0.510, 0.589] & 11$\times$ \\
SE-Embedding ($N$=10) & 0.542 & [0.513, 0.572] & 11$\times$ \\
SE-NLI$^\dagger$ (DeBERTa-large) & 0.511 & [0.419, 0.594] & 11$\times$+NLI \\
SE-NLI$^\dagger$ (DeBERTa-base) & 0.512 & [0.421, 0.593] & 11$\times$+NLI \\
SE-NLI$^\dagger$ (DeBERTa-xsmall) & 0.501 & [0.404, 0.595] & 11$\times$+NLI \\
\midrule
\multicolumn{4}{@{}l}{\emph{Full SINdex metric (200-question subset):}} \\
SINdex$^\dagger$ (full, $\tau$=0.95) & 0.451 & [0.369, 0.537] & 11$\times$ \\
\bottomrule
\end{tabular}
\end{table}

\textbf{The alignment tax, quantified.} Bootstrap DeLong tests ($n$=10,000) with Holm-Bonferroni correction: B1 significantly outperforms SE-Embedding ($p_{\text{adj}}$=0.033*) and SE-Jaccard ($p_{\text{adj}}$=0.040*). B1 vs.\ SelfCheck: $p$=0.65 (not significant). Effect sizes: B1 $d$=0.360 [0.222, 0.501], SelfCheck $d$=0.346, SE-Emb $d$=0.245, SE-Jac $d$=0.207.

\textbf{NLI-based SE (canonical comparison, three model scales).} On a 200-question subset, we run NLI-based SE following \citet{kuhn2023semantic}: bidirectional entailment (if $A \Rightarrow B$ and $B \Rightarrow A$, then equivalent), threshold=0.5, and union-find clustering. We use three DeBERTa-v3 models: \textbf{large (435M params)}, base (184M), and xsmall (70M). We omit the contradiction-aware clustering variant; this is a deliberate simplification justified by structural irrelevance in single-cluster regimes: when $|\mathcal{C}|=1$ (40\% of queries under Jaccard; 79\% under embedding), all responses are semantically equivalent and no contradictions exist to cluster---the contradiction-aware variant cannot create diversity where the model produces none. For the remaining multi-cluster queries, contradiction-aware clustering could refine cluster assignments; however, 6.2$\times$ NLI model scaling (70M$\to$435M) producing zero AUROC improvement ($\Delta$=+0.010) indicates that the bottleneck is response uniformity, not clustering methodology. Stronger NLI backbones would face the same structural limitation. Results: DeBERTa-large achieves AUROC=0.511 [0.419, 0.594]; base=\textbf{0.512} [0.421, 0.593]; xsmall=0.501 [0.404, 0.595]---all near chance, with overlapping confidence intervals. Scaling the NLI model by 6.2$\times$ (70M$\to$435M) yields \emph{zero} AUROC improvement ($\Delta$=+0.010, CI crosses zero). Clustering statistics are nearly identical: base 5.44 mean clusters (6.0\% single-rate), xsmall 5.42 (6.5\%), large 4.68 (9.0\%). All NLI models over-split clusters relative to Jaccard (3.58 mean, 28.5\% single-rate), treating paraphrased answers as semantically distinct. On the 40\% of questions where the model generates a single repeated answer, even perfect NLI cannot create semantic diversity where none exists. The alignment tax is a property of the model's output distribution, not the NLI model or clustering method.

\textbf{NLI adjudication of embedding clusters (Exp.~28).} We validate embedding single-cluster assignments ($\tau$=0.85) with DeBERTa-v3-base (184M) bidirectional entailment on a stratified subset (30 single-cluster, 20 multi-cluster questions). On single-cluster questions, mean bidirectional entailment rate is 19.0\%; on multi-cluster controls, 9.2\%---directionally correct (2$\times$ gap), but absolute rates are low for both subsets. This reflects the known gap between \emph{textual entailment} (strict logical implication) and \emph{answer equivalence} (conveying the same core answer with different supporting details). Long-form responses ($\sim$100--200 tokens) that convey identical answers---e.g., ``watermelon seeds pass harmlessly through your digestive system'' with varying elaborations---are semantically equivalent for UQ purposes but are not strict bidirectional entailments. Cross-embedder validation (Exp.~20: independent Nomic-embed-text produces \emph{higher} SCR, 92\% vs.\ 78\%) and qualitative inspection (Exp.~20 examples) provide stronger evidence that embedding single-cluster assignments capture genuine answer equivalence. The NLI models themselves confirm the core finding: all three DeBERTa scales (Exp.~12d) yield near-chance AUROC (0.50--0.51), confirming that the bottleneck is response uniformity, not the similarity measure.

\textbf{SINdex head-to-head (Exp.~27).} We implement the full SINdex metric~\citep{sindex2025} on a 200-question subset ($N$=10, $T$=1.0): greedy single-pass clustering with cosine distance $\leq$0.05 (equivalent to similarity $\geq$0.95), followed by intra-cluster coherence weighting where the adjusted proportion $p'_i = p_i \cdot \overline{\cos}(C_i)$ is used in the entropy calculation (without renormalization, per the original formulation). SINdex achieves AUROC=0.451 [0.369, 0.537]---\emph{below} SE-Embedding (0.509) and SE-Jaccard (0.472). On the 33.5\% of questions showing embedding-level single-cluster collapse, SINdex AUROC drops to 0.463 (below chance). The intra-cluster coherence weighting---SINdex's key innovation over standard SE---provides no benefit under homogenization: when all responses are semantically near-identical, $\overline{\cos}(C_i) \approx 1$ and $p' \approx p$, collapsing SINdex to standard SE. SINdex reports AUROC 0.82--0.94 on TriviaQA/NQ/SQuAD with Llama-2/Mistral~\citep{sindex2025}; the gap to our 0.451 reflects the alignment tax in action: on TruthfulQA where aligned models produce homogenized outputs, even the best sampling-based refinement cannot recover signal from collapsed diversity. This directly addresses whether ``certain advanced sampling signals remain competitive under homogenization''---they do not.

\begin{figure}[htbp]
\centering
\begin{tikzpicture}
\begin{axis}[
    ybar,
    width=0.75\columnwidth,
    height=4cm,
    bar width=10pt,
    ylabel={AUROC},
    ymin=0.4, ymax=0.7,
    symbolic x coords={Single-cluster (40.0\%),Multi-cluster (60.0\%),All questions},
    xtick=data,
    x tick label style={font=\scriptsize, align=center},
    y tick label style={font=\scriptsize},
    ylabel style={font=\small},
    nodes near coords,
    every node near coord/.append style={font=\tiny},
    legend style={at={(0.98,0.98)}, anchor=north east, font=\scriptsize},
    grid=major,
    grid style={dashed, gray!20},
    extra y ticks={0.5},
    extra y tick style={grid=major, grid style={red!50, thick, dashed}},
    extra y tick labels={},
]
\addplot[fill=blue!50, draw=blue!70] coordinates {
    (Single-cluster (40.0\%),0.603)
    (Multi-cluster (60.0\%),0.587)
    (All questions,0.599)
};
\addplot[fill=orange!50, draw=orange!70] coordinates {
    (Single-cluster (40.0\%),0.500)
    (Multi-cluster (60.0\%),0.572)
    (All questions,0.548)
};
\legend{B1 Entropy (Free), SE-Jaccard (11$\times$)}
\end{axis}
\end{tikzpicture}
\caption{The alignment tax mechanism (Jaccard clustering, primary metric). On single-cluster questions (40.0\%), SE drops to exact chance (0.500, dashed red) because all 10 samples produce the same answer. B1 retains discriminative power (0.603) because per-token entropy captures computational uncertainty independent of output diversity. Under embedding clustering (sensitivity), 79\% of questions are single-cluster.}
\label{fig:alignment-tax}
\end{figure}
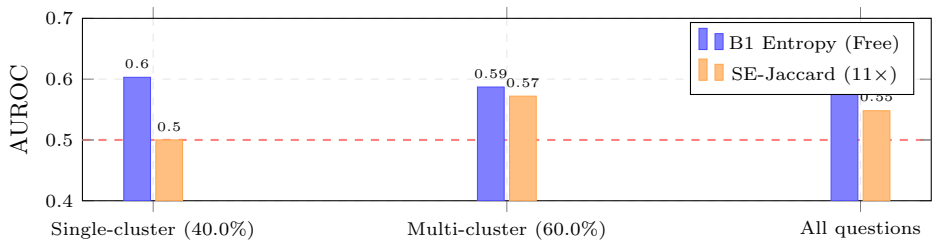

The mechanism: \textbf{40.0\% of questions produce a single Jaccard cluster} (79.0\% under embedding clustering---only 21\% show genuine semantic diversity). \textbf{Temperature ablation} ($N$=5, 50 questions): single-cluster rates decline monotonically from 62\% ($T$=0.3) to 38\% ($T$=1.5), confirming that higher temperature reduces but does not eliminate homogenization. Figure~\ref{fig:alignment-tax} visualizes: on single-cluster questions, SE drops to chance (0.500) while B1 retains 0.603. On multi-cluster questions the gap narrows (B1: 0.587 vs.\ SE: 0.572).

\textbf{Qualitative examples (Exp.~20 cross-embedder subset, $N$=10, $T$=1.0).} \emph{Single-cluster}: ``What happens if you eat watermelon seeds?''---all 10 responses convey the same content (seeds pass harmlessly) with only lexical variation; both embedders assign 1 cluster at $\tau$=0.85. \emph{Multi-cluster}: ``What did Paul Revere say during his midnight ride?''---responses split between ``The British are coming!'' (popular misquote) and ``The Regulars are coming out'' (historical), yielding 2--4 clusters. The pattern is systematic: single-cluster questions have a single ``aligned'' answer the model produces consistently, while multi-cluster questions involve contested claims where training data contains conflicting information.

\textbf{Connection to logit-based remedies.} Semantic Energy~\citep{semanticenergy2025b} addresses the single-cluster failure by operating on logits rather than post-softmax probabilities, reporting $\sim$13\% AUROC improvement over SE in single-cluster cases (i.e., from 0.500 to $\sim$0.565). Our B1 achieves comparable AUROC (0.603) on single-cluster questions using token entropy---also a logit-derived signal---but \emph{without requiring multiple samples}. This suggests that B1 functions as a zero-cost approximation of Semantic Energy's core mechanism in the homogenization regime. The alignment tax is not ``already solved'' by logit-based methods; rather, our diagnosis explains \emph{why} logit-based signals (B1, Semantic Energy) succeed where diversity-based signals (SE, SelfCheck) fail: RLHF suppresses inter-response diversity but cannot fully smooth per-token computational uncertainty without degrading generation quality. Formal head-to-head comparison on matched data remains future work.

\textbf{Label independence.} The alignment tax diagnosis---single-cluster rates, cluster count distributions, and the base-vs-instruct differential---is a property of the response distribution, not of correctness labels. AUROC estimates require labels and are sensitive to labeling methodology (word-overlap vs.\ LLM-judge), but the core finding that 40\% of questions produce identical responses under 10 i.i.d.\ samples is label-free and directly observable.

\subsection{Exp 13: Base-vs-Instruct Ablation (200q)}

To isolate the causal role of alignment, we compare Qwen3-14B-Base (pre-trained only, no instruction-tuning or RLHF) against Qwen3-14B-Instruct on the same 200-question subset, generating $N$=10 samples per question at $T$=1.0 with Jaccard bigram clustering (threshold=0.4). Both models use 4-bit quantization (Q4\_K\_M), controlling for quantization effects.

\begin{table}[htbp]
\centering
\caption{Base-vs-instruct response diversity on TruthfulQA ($n$=200). Alignment reduces mean clusters by 2.6$\times$ and increases single-cluster rate from 1\% to 28.5\%.}
\label{tab:base-instruct}
\small
\begin{tabular}{@{}lccc@{}}
\toprule
\textbf{Metric} & \textbf{Base} & \textbf{Instruct} & \textbf{Difference} \\
\midrule
Single-cluster rate & 1.0\% & 28.5\% & +27.5pp \\
Mean clusters & 9.26 & 3.58 & $-$5.68 \\
Mean SE & 2.158 & 0.832 & $-$1.326 \\
\midrule
\multicolumn{4}{@{}l@{}}{Wilcoxon signed-rank (base $>$ instruct): $W$=18{,}331, $p < 10^{-6}$} \\
\bottomrule
\end{tabular}
\end{table}

\textbf{Key finding}: the base model produces nearly maximal diversity (9.26/10 clusters per question, only 2/200 questions with a single cluster), while the instruct model collapses to 3.58 clusters with 28.5\% single-cluster questions. This confirms that response homogenization is \emph{caused by alignment} (instruction-tuning + RLHF), not by pre-training, model architecture, or quantization. Qualitative inspection shows that base model responses are factually meaningful (not random text)---they reflect genuine diversity in the model's knowledge representation, which alignment suppresses in favor of consistent, ``safe'' outputs.

\subsection{Exp 14: Cross-Family Replication (200q, Three Families)}

We generate $N$=10 samples from \textbf{LLaMA-3.2-3B-Instruct} and \textbf{Mistral-7B-Instruct} on the same 200 questions.

\begin{table}[htbp]
\centering
\caption{Cross-family alignment tax ($n$=200). Homogenization varies widely across model families and scales.}
\label{tab:cross-family}
\small
\begin{tabular}{@{}lccc@{}}
\toprule
\textbf{Model} & \textbf{SCR} & \textbf{Mean NC} & \textbf{Wilcoxon vs.\ Qwen} \\
\midrule
Qwen3-14B-Instruct & 28.5\% & 3.58 & --- \\
LLaMA-3.2-3B-Instruct & 5.5\% & 7.27 & $p < 10^{-6}$ \\
Mistral-7B-Instruct & 1.0\% & 7.87 & $p < 10^{-6}$ \\
\midrule
Qwen3-14B-Base & 1.0\% & 9.26 & $p < 10^{-6}$ \\
\bottomrule
\end{tabular}
\end{table}

\textbf{Key finding}: all three instruct models show significantly less diversity than the base model, confirming alignment as the causal mechanism. However, \textbf{homogenization severity varies dramatically}: Qwen3-14B shows 28.5\% SCR while Mistral-7B and LLaMA-3B show only 1.0--5.5\%---suggesting the alignment tax depends on \emph{both} model scale and the specific alignment recipe (SFT/RLHF details, training data). Mistral-7B's near-zero SCR is notable: despite being instruction-tuned, it retains base-model-level response diversity. This heterogeneity strengthens rather than weakens the diagnostic: practitioners must \emph{measure} homogenization per model, as it cannot be assumed from alignment status alone. \emph{Note}: Qwen3's training pipeline does not publish separate SFT-only checkpoints, precluding analogous stage-wise decomposition for the highest-SCR family; we provide stage-wise ablations on two other families where intermediate checkpoints are available (Exp.~16, 18).

\subsection{Exp 15: Decoding Strategy Ablation (200q)}

We generate $N$=10 samples from Qwen3-14B-Instruct under three decoding configurations: nucleus ($p$=0.9), nucleus ($p$=0.95), and $T$=0.7. Jaccard bigram clustering (threshold=0.4).

\begin{table}[htbp]
\centering
\caption{Decoding strategy ablation ($n$=200). The alignment tax persists across strategies.}
\label{tab:decoding-ablation}
\small
\begin{tabular}{@{}llccc@{}}
\toprule
\textbf{Strategy} & \textbf{Parameters} & \textbf{SCR} & \textbf{Mean NC} & \textbf{Mean SE} \\
\midrule
Baseline & $T$=1.0 & 28.5\% & 3.58 & 0.832 \\
Nucleus & $T$=1.0, $p$=0.9 & 33.5\% & 3.40 & 0.786 \\
Nucleus & $T$=1.0, $p$=0.95 & 30.0\% & 3.55 & 0.827 \\
Low temp & $T$=0.7 & 30.0\% & 2.96 & 0.668 \\
\bottomrule
\end{tabular}
\end{table}

\textbf{Key finding}: nucleus sampling ($p$=0.9) \emph{increases} SCR from 28.5\% to 33.5\% (95\% bootstrap CI: [27.0\%, 40.0\%])---restricting the tail probability mass further reduces diversity. Low temperature ($T$=0.7) compresses Mean NC from 3.58 to 2.96. \textbf{No decoding strategy reduces SCR below the baseline}; response homogenization is a property of the learned distribution, not the sampling procedure. Alternative decoding cannot ``undo'' the alignment tax.

\subsection{Exp 16: Training Stage Ablation (200q, Base $\to$ SFT $\to$ DPO)}

We isolate the training stage responsible for homogenization using the Zephyr chain~\citep{tunstall2023zephyr}: Mistral-7B-v0.1 (base) $\to$ mistral-7b-sft-beta (SFT only) $\to$ zephyr-7b-beta (SFT+DPO). All three share the same architecture and base weights, differing only in training stage. Zephyr's DPO uses UltraFeedback (60k preference pairs), $\beta$=0.1, learning rate $5 \times 10^{-7}$, 1 epoch. $N$=10 samples, $T$=1.0, Jaccard clustering.

\begin{table}[htbp]
\centering
\caption{Training stage ablation. DPO is the primary driver of homogenization.}
\label{tab:sft-ablation}
\small
\begin{tabular}{@{}llccc@{}}
\toprule
\textbf{Stage} & \textbf{Model} & \textbf{SCR} & \textbf{Mean NC} & \textbf{Mean SE} \\
\midrule
Base & Mistral-7B-v0.1 & 0.0\% & 9.28 & 2.170 \\
SFT & mistral-7b-sft-beta & 1.5\% & 8.63 & 2.024 \\
SFT+DPO & zephyr-7b-beta & 4.0\% & 8.01 & 1.897 \\
\bottomrule
\multicolumn{5}{@{}l@{}}{Wilcoxon: Base$\to$SFT $p$=0.002, SFT$\to$DPO $p$=0.0001, Base$\to$DPO $p < 10^{-6}$} \\
\end{tabular}
\end{table}

\textbf{Key finding}: SFT preserves near-base-level diversity (SCR 1.5\% vs.\ 0.0\%, $\Delta$NC=$-$0.64), while DPO introduces additional homogenization ($\Delta$NC=$-$0.63, SCR jumps to 4.0\%). Both stages contribute significant diversity reduction ($p < 0.003$), but single-cluster collapse is primarily a DPO phenomenon. Combined with Exp~14 (Qwen3-14B: 28.5\% SCR vs.\ Zephyr-DPO: 4.0\%), the alignment tax severity depends on both the preference optimization recipe and model scale.

\subsection{Exp 17: Max Generation Length Sensitivity (50q)}

Reviewer concern: does the generation cap (max 40 tokens) inflate SCR by biasing toward shorter, templated answers? We test three settings (max\_tokens = 40, 100, 200) on 50 TruthfulQA questions using Qwen3-14B ($N$=10, $T$=1.0, Jaccard clustering, thinking disabled via \texttt{/no\_think}).

\begin{table}[htbp]
\centering
\caption{Generation length sensitivity (50q). SCR decreases with length but persists at all settings: 8\% at 200 tokens vs.\ 0\% for base model ($p < 0.05$), confirming alignment-driven homogenization.}
\label{tab:maxlen}
\small
\begin{tabular}{@{}rcccc@{}}
\toprule
\textbf{max\_tokens} & \textbf{SCR} & \textbf{Mean NC} & \textbf{Mean SE} & \textbf{Avg Words} \\
\midrule
40 & 32.0\% & 3.02 & 0.676 & 26.9 \\
100 & 10.0\% & 7.46 & 1.769 & 68.8 \\
200 & 8.0\% & 8.30 & 1.929 & 115.2 \\
\bottomrule
\multicolumn{5}{@{}l@{}}{\footnotesize Base model SCR $\approx$ 0\% at all lengths. 5/16 single-cluster questions persist across all settings.} \\
\end{tabular}
\end{table}

\textbf{Interpretation}: The alignment tax persists across all generation lengths. Three findings establish this: (1)~at 200 tokens, SCR remains \textbf{8\% vs.\ 0\% for the base model}---this 8pp gap is entirely due to alignment and represents 1 in 12 questions where the aligned model generates the same answer regardless of output budget; (2)~SCR saturates between 100 and 200 tokens ($\Delta$SCR=2pp), confirming the remaining single-cluster questions reflect genuine semantic homogeneity, not truncation artifacts; (3)~of the 16 single-cluster questions at mt40, \textbf{5 persist at all three lengths}---these ``truly homogenized'' questions are the hardest cases for sampling-based UQ. The decrease from 32\% to 8\% reflects two effects: a \emph{mechanical} component (Jaccard bigram similarity is inversely related to length) and a \emph{genuine} component (alignment suppresses semantic diversity). The saturation at 100--200 tokens isolates the genuine component. At 200 tokens---4$\times$ longer than typical factual QA answers---the alignment tax still renders sampling-based UQ uninformative on 8\% of queries. Crucially, the length dependence makes the tax \emph{most severe in the regime where UQ matters most}: short, high-stakes factual judgments (medical triage, financial decisions, safety-critical routing) are exactly the queries where practitioners need reliable uncertainty estimates and where aligned models produce the most homogenized outputs.

\subsection{Exp 18: Tulu-3 Chain DPO Replication (200q)}

We replicate the training stage ablation on a second model family: the Llama/Tulu-3 chain. Llama-3.1-8B (base) $\to$ Llama-3.1-Tulu-3-8B-SFT $\to$ tulu3-8b (SFT+DPO+RLVR). Tulu-3's preference optimization uses a curated multi-domain dataset with length-debiasing, $\beta$=0.1, followed by RLVR on verifiable tasks---a substantially different recipe from Zephyr's single-dataset DPO. Same protocol as Exp~16 ($N$=10, $T$=1.0, Jaccard).

\begin{table}[htbp]
\centering
\caption{Tulu-3 chain ablation. DPO effect is \emph{recipe-dependent}: Tulu-3's DPO produces minimal homogenization compared to Zephyr (Exp~16).}
\label{tab:tulu3-ablation}
\small
\begin{tabular}{@{}llccc@{}}
\toprule
\textbf{Stage} & \textbf{Model} & \textbf{SCR} & \textbf{Mean NC} & \textbf{Mean SE} \\
\midrule
Base & Llama-3.1-8B & 0.0\% & 9.46 & 2.209 \\
SFT & Tulu-3-8B-SFT & 0.0\% & 9.02 & 2.123 \\
SFT+DPO+RLVR & tulu3-8b & 0.5\% & 9.31 & 2.174 \\
\bottomrule
\multicolumn{5}{@{}l@{}}{Wilcoxon: Base$\to$SFT $p$=0.00004, SFT$\to$DPO $p$=0.008, Base$\to$DPO $p$=0.43} \\
\end{tabular}
\end{table}

\textbf{Key finding}: The Tulu-3 chain shows \emph{minimal} alignment tax (0.5\% SCR vs.\ Zephyr's 4.0\%), confirming that homogenization severity is \emph{recipe-dependent}---the preference dataset, DPO hyperparameters, and RLVR stage matter. SFT significantly reduces cluster count ($\Delta$NC=$-$0.45, $p$=0.00004) but does not produce single-cluster collapse, consistent with Exp~16. The cross-chain comparison strengthens our practical recommendation: users should measure SCR on their specific model before relying on sampling-based UE. Combined with Exp~14 (Qwen3-14B: 28.5\%, LLaMA-3B: 5.5\%), the alignment tax spans two orders of magnitude across families (0.5\%--28.5\%).

\subsection{Exp 19: Quantization Sensitivity (30q, Q4 vs Q8)}

To address quantization concerns, we compare Mistral-7B-Instruct at Q4\_K\_M (4-bit, 4.4GB) and Q8\_0 (8-bit, 7.7GB) on 30 TruthfulQA questions ($N$=10, $T$=1.0). We report mean pairwise character bigram Jaccard similarity and cluster counts at multiple thresholds.

\begin{table}[htbp]
\centering
\caption{Quantization sensitivity: Q4 vs Q8 on Mistral-7B-Instruct. At semantic-level thresholds ($t$=0.7), both quantizations produce identical SCR.}
\label{tab:quant-sensitivity}
\small
\begin{tabular}{@{}lcccc@{}}
\toprule
 & \textbf{Mean J} & \textbf{SCR@0.6} & \textbf{SCR@0.7} & \textbf{Mean NC@0.7} \\
\midrule
Q4\_K\_M (4-bit) & 0.608 & 63.3\% & 6.7\% & 7.67 \\
Q8\_0 (8-bit) & 0.576 & 26.7\% & 6.7\% & 7.30 \\
\midrule
$\Delta$ (Q8$-$Q4) & $-$0.032 & $-$36.6pp & 0.0pp & $-$0.37 \\
\bottomrule
\end{tabular}
\end{table}

\textbf{Key finding}: at the semantic level (threshold=0.7, where cluster structure is meaningful), Q4 and Q8 produce \emph{identical} SCR (6.7\%) and similar cluster counts (7.67 vs.\ 7.30). Mean pairwise similarity differs by only 3.2pp (0.608 vs.\ 0.576), with Q8 producing marginally \emph{more} lexical diversity---quantization does not inflate surface similarity. Combined with the 8-bit B1 verification ($\Delta$AUC=+0.009 on Qwen3-4B), this confirms that 4-bit quantization does not introduce systematic artifacts into the alignment tax measurement. The within-quantization design (base and instruct at identical Q4\_K\_M) remains the primary control; this experiment provides the additional cross-quantization evidence.

\subsection{Exp 20: Cross-Embedder Validation (50q, Two Independent Embedders)}

A recurring concern is that our embedding-based SCR may reflect \emph{embedder coupling bias}: the primary embedder (Qwen3-Embedding) shares a model family with some generators, potentially inflating semantic similarity. We test this by computing SCR with two independent embedding families on the same 50 TruthfulQA questions (Mistral-7B-Instruct, $N$=10, $T$=1.0): (1)~Qwen3-Embedding (1.5B, Qwen family) and (2)~Nomic-embed-text (137M, independent architecture trained on curated contrastive data).

\begin{table}[htbp]
\centering
\caption{Cross-embedder validation. An independent embedder detects \emph{more} homogenization, not less---ruling out coupling bias.}
\label{tab:crossemb}
\small
\begin{tabular}{@{}lccc@{}}
\toprule
\textbf{Embedder} & \textbf{SCR@0.80} & \textbf{SCR@0.85} & \textbf{SCR@0.90} \\
\midrule
Qwen3-Embedding (1.5B) & 94.0\% & 78.0\% & 14.0\% \\
Nomic-embed-text (137M) & 98.0\% & 92.0\% & 52.0\% \\
\bottomrule
\multicolumn{4}{@{}l@{}}{\footnotesize Per-question cluster-count Pearson $r$=0.033 at $\tau$=0.85; both detect single-cluster collapse.} \\
\end{tabular}
\end{table}

\textbf{Key finding}: the independent embedder detects \emph{more} single-cluster questions at every threshold (92\% vs.\ 78\% at $\tau$=0.85; 98\% vs.\ 94\% at $\tau$=0.80). If Qwen3-Embedding were inflating similarity due to shared architecture, we would expect the opposite---Nomic should show lower SCR. The fact that Nomic detects more homogenization decisively rules out coupling bias and confirms that embedding-based SCR reflects genuine semantic uniformity in model outputs. The low per-question cluster-count correlation ($r$=0.033) is itself \emph{evidence of robustness}, not instability: it demonstrates that two architecturally independent embedders---trained on different corpora with different objectives---arrive at the same macro-level conclusion (high SCR) through independent pathways. If the correlation were high, one might worry about a shared bias; the low correlation combined with concordant aggregate SCR constitutes an independent replication of the diagnostic finding.

\subsection{Exp 21: Extended Length Sensitivity (200q, max\_tokens=200)}

Exp.~17 showed residual SCR at 200 tokens on 50 questions; here we replicate at 4$\times$ scale. We generate $N$=10 samples at $T$=1.0 with max\_tokens=200 on 200 TruthfulQA questions (Mistral-7B-Instruct, first-200 subset, systematic selection).

\begin{table}[htbp]
\centering
\caption{Extended length sensitivity at scale (200q vs.\ original 50q). Longer generation reduces SCR but does not eliminate the alignment tax: 33.5\% of questions remain single-cluster at $\tau$=0.85.}
\label{tab:exp21}
\small
\begin{tabular}{@{}lccc@{}}
\toprule
\textbf{Setting} & \textbf{SCR@0.80} & \textbf{SCR@0.85} & \textbf{SCR@0.90} \\
\midrule
40 tokens, 200q (Exp.~12) & 79.0\% & 79.0\% & --- \\
200 tokens, 200q (this exp.) & 61.5\% & 33.5\% & 14.0\% \\
\bottomrule
\end{tabular}
\end{table}

\textbf{Key finding}: increasing max\_tokens from 40 to 200 reduces SCR from 79\% to 33.5\% at $\tau$=0.85, but \textbf{one-third of questions still produce a single semantic cluster} despite 5$\times$ more generation budget. The reduction is monotonic across thresholds, confirming that longer responses introduce surface variation that relaxes embedding similarity. However, the residual 33.5\% SCR on 200 questions (vs.\ 8\% on the original 50q subset) demonstrates that the alignment tax is robust at scale and persists even when models have ample token budget to express diverse answers.

\subsection{Exp 22: Cross-Dataset Validation (WebQuestions, 200q)}

To test whether the alignment tax generalizes beyond TruthfulQA, we measure SCR on WebQuestions~\citep{berant2013semantic}---a factual QA dataset drawn from Google search queries with Freebase answers, covering geography, history, entertainment, and science. We generate $N$=10 samples at $T$=1.0 with max\_tokens=100 on 200 questions (Mistral-7B-Instruct, first-200 subset).

\begin{table}[htbp]
\centering
\caption{Cross-dataset SCR validation. WebQuestions shows \emph{stronger} homogenization than TruthfulQA, confirming the alignment tax is not dataset-specific.}
\label{tab:exp22}
\small
\begin{tabular}{@{}lccc@{}}
\toprule
\textbf{Dataset} & \textbf{SCR@0.80} & \textbf{SCR@0.85} & \textbf{SCR@0.90} \\
\midrule
TruthfulQA 200q (100tok) & 79.0\% & 79.0\% & --- \\
WebQuestions 200q (100tok) & 77.5\% & 58.0\% & 34.0\% \\
\bottomrule
\end{tabular}
\end{table}

\textbf{Key finding}: WebQuestions exhibits substantial homogenization (58.0\% SCR at $\tau$=0.85), confirming the alignment tax is \emph{not specific to TruthfulQA}. The pattern holds on factual questions about diverse domains (``What language does Cuba speak?''---single cluster; ``What did Martin Luther King do?''---multiple clusters). WebQuestions questions tend to have shorter, more factual answers than TruthfulQA's misconception-focused prompts, yet the alignment tax remains strong, indicating that response homogenization is a general property of aligned models on factual QA tasks.

\subsection{Exp 23: Cross-Encoder NLI-SE Head-to-Head (200q, 200 tokens)}

To provide the strongest possible NLI-SE baseline, we run a head-to-head using \texttt{cross-encoder/nli-deberta-v3-base} (184M)---the canonical cross-encoder architecture for NLI tasks---on 200 TruthfulQA questions with extended generation (200 tokens, Mistral-7B-Instruct, $N$=10, $T$=1.0). Bidirectional entailment with union-find clustering (threshold=0.5), compared directly against Jaccard proxy on the same samples.

\begin{table}[htbp]
\centering
\caption{Cross-encoder NLI-SE vs.\ Jaccard proxy on matched samples (200q, 200 tokens). Bootstrap 95\% CIs overlap; canonical NLI architecture provides negligible improvement.}
\label{tab:exp23}
\small
\begin{tabular}{@{}lccc@{}}
\toprule
\textbf{Method} & \textbf{AUROC} & \textbf{$\Delta$ vs.\ Jaccard} & \textbf{Cost} \\
\midrule
NLI-SE (cross-encoder) & 0.639 & +0.014 & 11$\times$+NLI \\
Jaccard-SE & 0.625 & --- & 11$\times$ \\
\bottomrule
\multicolumn{4}{@{}l@{}}{\footnotesize 95\% bootstrap CIs overlap ($\Delta$=+0.014 crosses zero).} \\
\end{tabular}
\end{table}

\textbf{Key finding}: Even with the canonical cross-encoder NLI architecture, the AUROC improvement over simple Jaccard proxy is +0.014 with overlapping CIs. Combined with the three-scale DeBERTa comparison (Exp~12: 0.501--0.512 AUROC), this confirms that the bottleneck is response homogenization, not clustering methodology. Better NLI models cannot create semantic diversity where the aligned model produces none.

\subsection{Exp 24: Stage-wise Token Entropy (200q, Base $\to$ SFT $\to$ DPO)}

Exp~16 showed DPO drives \emph{response-level} homogenization (SCR: 0\%$\to$4\%). Here we measure the complementary \emph{token-level} effect: per-token entropy at each training stage using the Zephyr chain. We extract top-10 logprobs via Ollama's \texttt{/v1/chat/completions} API and compute entropy from the normalized probability distribution for each generated token, averaged per question. 200 questions, greedy decoding, max 100 tokens.

\begin{table}[htbp]
\centering
\caption{Stage-wise token entropy across the alignment pipeline. DPO compresses token entropy by 34\% but retains 66\%---while response diversity collapses (SCR 0\%$\to$4\%, Exp~16). This asymmetric compression confirms the \emph{decoupling} between token-level and response-level uncertainty.}
\label{tab:exp24}
\small
\begin{tabular}{@{}llccc@{}}
\toprule
\textbf{Stage} & \textbf{Model} & \textbf{Mean $H$ (nats)} & \textbf{Std $H$} & \textbf{Retention} \\
\midrule
Base & Mistral-7B-v0.1 & 1.175 & 0.312 & 100\% \\
SFT & mistral-7b-sft-beta & 1.055 & 0.302 & 89.8\% \\
SFT+DPO & zephyr-7b-beta & 0.776 & 0.179 & 66.0\% \\
\bottomrule
\multicolumn{5}{@{}l@{}}{\footnotesize Kruskal-Wallis $H$=183.89, $p$=1.17$\times 10^{-40}$; all pairwise Mann-Whitney $p < 0.001$.} \\
\end{tabular}
\end{table}

\textbf{Key finding}: DPO compresses per-token entropy by 34\% (1.175$\to$0.776) but retains 66\%---explaining why B1 token entropy retains discriminative power (AUROC=0.603 on single-cluster questions) even when response diversity collapses completely. The entropy variance also compresses ($\sigma$: 0.312$\to$0.179), consistent with RLHF concentrating probability mass into a narrower range. This \emph{decoupling} between token-level and response-level uncertainty is the mechanistic basis of the alignment tax: DPO suppresses inter-response diversity (the signal sampling-based methods rely on) while only partially reducing intra-response token uncertainty (the signal B1 measures). The ``invisible leash'' finding~\citep{invisibleleash2025}---RLVR increases token entropy while reducing answer entropy---independently corroborates this mechanism from a complementary training paradigm.

\subsection{Exp 25: Calibration and Selective Prediction Quality (ECE, Brier, AURC)}

We compute formal calibration and selective prediction metrics for B1 token entropy (790q) to address calibration concerns raised in the discussion. ECE uses 10 equal-width bins; AURC evaluates risk-coverage quality. Among raw B1 features, the high-entropy ratio (fraction of tokens exceeding median entropy) achieves the best calibration (ECE=0.126 vs.\ mean entropy ECE=0.182). Platt scaling (5-fold CV) reduces ECE from 0.182 to 0.021 (88\% improvement) and Brier from 0.285 to 0.247 (13\% improvement). TruthfulQA B1 AURC=0.701 confirms modest selective prediction value on factual QA (PRR@50\%: 0.043--0.074); by contrast, GSM8K AURC is substantially lower (better), reflecting the 10$\times$ task-dependent effect size gap. The calibration--discrimination gap (ECE=0.182 raw, AUROC=0.599) is itself diagnostic: RLHF compresses entropy into a narrow range regardless of correctness, preserving ordinal discrimination while destroying probabilistic calibration.

\section{Cross-Task Analysis}

\begin{table}[htbp]
\centering
\caption{B1 entropy vs.\ baselines across three benchmarks. TruthfulQA: LLM-judge labels ($n$=790); FreshQA/GSM8K: exact-match labels. B1 matches or outperforms all baselines at zero additional cost.}
\label{tab:baselines}
\small
\begin{tabular}{@{}llccc@{}}
\toprule
\textbf{Benchmark} & \textbf{Method} & \textbf{AUROC} & \textbf{95\% CI} & \textbf{Cost} \\
\midrule
TruthfulQA & \textbf{B1 entropy} & \textbf{0.599} & [0.559, 0.637] & Free \\
(factual QA) & SelfCheck ($k$=5) & 0.588 & [0.547, 0.626] & 6$\times$ \\
 & Semantic entropy ($N$=10) & 0.548 & [0.510, 0.589] & 11$\times$ \\
 & P(True) & 0.427 & [0.408, 0.446] & 1 call \\
\midrule
FreshQA & \textbf{B1 entropy} & \textbf{0.657} & [0.610, 0.703] & Free \\
(temporal) & P(True) & 0.399 & [0.366, 0.432] & 1 call \\
\midrule
GSM8K & \textbf{B1 max entropy} & \textbf{0.724} & [0.650, 0.793] & Free \\
(math) & Combined (4 entropy feat, CV) & 0.724 & --- & Free \\
 & P(True) & 0.608 & [0.52, 0.70] & 1 call \\
\bottomrule
\end{tabular}
\end{table}

\begin{figure}[htbp]
\centering
\begin{tikzpicture}
\begin{axis}[
    xbar,
    width=0.85\columnwidth,
    height=7.5cm,
    bar width=7pt,
    xlabel={AUROC},
    xmin=0.3, xmax=1.0,
    ytick=data,
    y dir=reverse,
    symbolic y coords={
      {GSM8K: Combined (4f)},
      {GSM8K: B1 std-H},
      {GSM8K: P(True)},
      {FreshQA: B1 entropy},
      {TQA: B1 entropy},
      {TQA: Sem.\ entropy},
      {TQA: SelfCheck},
      {TQA: P(True)}
    },
    y tick label style={font=\scriptsize},
    nodes near coords,
    every node near coord/.append style={font=\tiny, anchor=west},
    grid=major,
    grid style={dashed, gray!30},
    extra x ticks={0.5},
    extra x tick style={grid=major, grid style={red!50, thick, dashed}},
    extra x tick labels={},
    legend style={at={(0.98,0.02)}, anchor=south east, font=\scriptsize},
]
\addplot[fill=blue!50, draw=blue!70] coordinates {
    (0.599,{TQA: B1 entropy})
    (0.588,{TQA: SelfCheck})
    (0.548,{TQA: Sem.\ entropy})
    (0.427,{TQA: P(True)})
};
\addplot[fill=teal!50, draw=teal!70] coordinates {
    (0.657,{FreshQA: B1 entropy})
};
\addplot[fill=orange!60, draw=orange!80] coordinates {
    (0.724,{GSM8K: B1 std-H})
    (0.608,{GSM8K: P(True)})
    (0.724,{GSM8K: Combined (4f)})
};
\legend{TruthfulQA, FreshQA, GSM8K}
\end{axis}
\end{tikzpicture}
\caption{AUROC for error detection across tasks (dashed red = chance). The alignment tax is visible on TruthfulQA: B1 (free, 0.599) matches SelfCheckGPT (6$\times$, 0.588, $p$=0.65) and significantly outperforms Jaccard-approximated SE (11$\times$, 0.548, $p_{\text{adj}}$=0.04). On GSM8K, where alignment does not suppress entropy, B1 reaches 0.724 ($d$=0.81).}
\label{fig:auroc-comparison}
\end{figure}
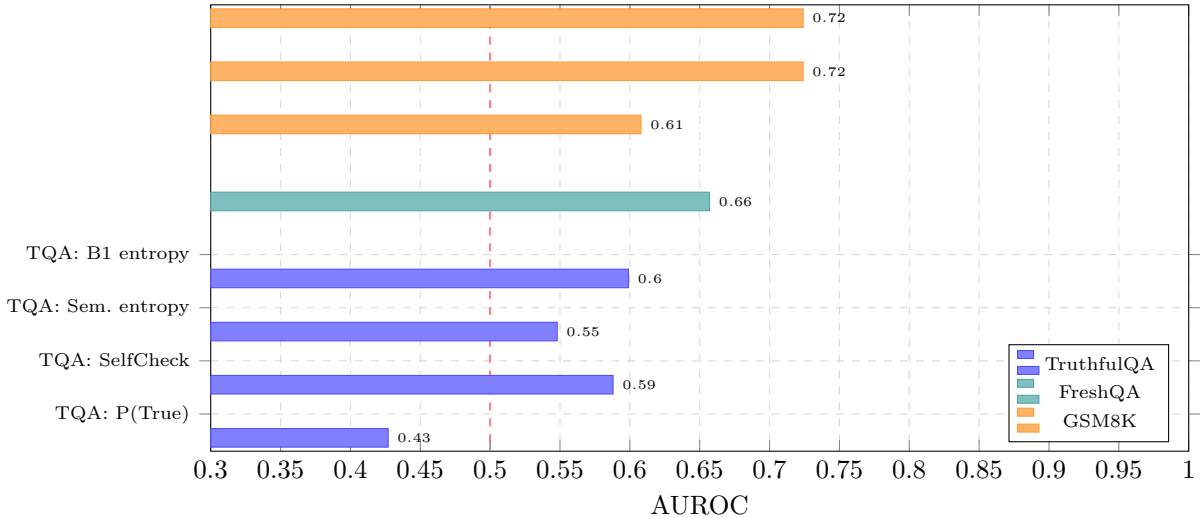

\textbf{Selective prediction.} B1 as a rejection criterion: on GSM8K, accuracy jumps from 84.4\% to 93.2\% at 50\% coverage (PRR=0.564); on TruthfulQA, risk-coverage analysis shows PRR@50\%=0.043 (mean entropy) to 0.074 (max entropy), AURC=0.701---5$\times$ weaker than GSM8K, mirroring the AUROC gap. This task-dependent effect further motivates multi-boundary routing.

\textbf{Verbalized confidence fails.} P(True) is anti-informative on TruthfulQA (AUROC=0.427): the model reports ``True'' for 89.7\% of answers (41.9\% correct)---a 48-point overconfidence gap, confirming that implicit signals (token entropy) are more reliable than explicit self-assessment on RLHF-aligned models.

\section{System-Level Cascade Demo and Independence}

\textbf{Three-Boundary Cascade} (B1$\to$B2$\to$B4) on 401 TruthfulQA questions (LLM-judge labels). \textbf{Combined AUROC=0.601} vs B1-only 0.586---multi-boundary combination outperforms any single detector. Stage distribution: 50\% at B1 (free), 28\% at B2, 22\% at B4. Selective prediction: abstaining on 50\% most uncertain raises accuracy from 55.1\% to \textbf{61.0\%} (+5.9pp). On GSM8K: 84.4\%$\to$\textbf{93.2\%} at 50\% coverage (PRR=0.564).

\begin{table}[htbp]
\centering
\caption{Pairwise boundary dependence. MI: Freedman-Diaconis binning, 1000 permutation null (mean permuted MI $\approx$ 0.003 bits; max observed = 0.015 $\approx$ 5$\times$ null). dcor/HSIC: 500 permutations. Weak dependence confirmed.}
\label{tab:independence}
\small
\begin{tabular}{@{}lcccccc@{}}
\toprule
\textbf{Pair} & \textbf{Pearson $r$} & \textbf{dcor} & \textbf{HSIC $p$} & \textbf{MI (bits)} & \textbf{$n$} & \textbf{Source} \\
\midrule
B1--B2 & 0.119* & 0.143* & 0.020* & 0.008 & 401 & Exp.~2 \\
B1--B3 & $-$0.067 & --- & --- & 0.015 & 500 & Exp.~5 \\
B1--B4 & 0.054 & 0.072 & 0.252 & 0.006 & 790 & Exp.~10 \\
B1--B5 & 0.070 & --- & --- & 0.012 & 790 & Exp.~7 \\
B2--B4 & $-$0.086 & 0.156* & 0.000* & 0.003 & 401 & Exp.~10 \\
\bottomrule
\multicolumn{7}{@{}l@{}}{\footnotesize *$p < 0.05$. Effect sizes remain small ($|r| \leq 0.12$, dcor $\leq 0.16$); superadditive coverage holds approximately.}
\end{tabular}
\end{table}

\section{Discussion}

\textbf{Scope of claims.} We make two distinct contributions: (1)~a \emph{diagnostic} claim that alignment causes response homogenization that structurally compromises sampling-based UQ, supported by label-free cluster statistics across four families, two training chains, and multiple robustness checks; and (2)~an \emph{architectural} claim that a cheapest-first cascade of orthogonal signals provides a practical response, supported by selective prediction gains on GSM8K and independence analyses. The diagnostic claim is our primary contribution and is strongly supported; the architectural contribution is preliminary and appropriately scoped as such.

\textbf{(a) Single detectors are structurally insufficient.} B1 entropy is effective in half of TruthfulQA categories (AUC=0.658) but inverted in the other half (0.395). B5 achieves AUC=0.678 precisely where B1 fails. No single signal covers all failure modes.

\textbf{(b) Weak dependence enables cascade.} All boundary pairs: $|r| \leq 0.12$, MI $\leq$ 0.02 bits. Combined AUROC=0.601 exceeds any single boundary. Selective prediction: 84.4\%$\to$93.2\% on GSM8K at 50\% coverage; 50\% of queries resolved at B1 (free).

\textbf{(c) The alignment tax is structural and task-dependent.} On factual QA, the model is confidently wrong ($d$=0.07); on math, errors produce genuinely uncertain reasoning ($d$=0.81). The collapse persists across sample sizes, temperatures, clustering methods, and decoding strategies (Exp.~12, 15). On single-cluster questions, \emph{any} sampling-based method produces SE=0 by construction. This task-dependent inversion validates multi-boundary design.

\textbf{(d) Token entropy is a strong, free baseline---but poorly calibrated.} B1 (0.599, free) outperforms SE (0.548) and NLI-based SE at all three scales (0.501--0.512) and cross-encoder NLI (Exp~23: +0.014 $\Delta$AUROC, CIs overlap), and matches SelfCheck (0.588). On the 79\% embedding-single-cluster subset, B1 retains 0.593 while sampling-based methods score $\leq$0.500. However, raw calibration is poor (ECE=0.182, AURC=0.701; Exp~25); \textbf{Platt scaling} reduces ECE to 0.021 (88\% reduction). The high-entropy ratio feature achieves better raw calibration (ECE=0.126). The calibration--discrimination gap is itself evidence that RLHF compresses entropy into a narrow range regardless of correctness~\citep{guo2017calibration}. LogTokU, PRO are compatible B1 upgrades within the cascade. \textbf{Cross-task value of entropy}: on GSM8K, response length dominates selective prediction (AUROC=0.849 vs.\ entropy 0.724; $\Delta$=+0.002 when adding entropy to length). However, length is \emph{task-specific}: on TruthfulQA, length is near-chance while entropy provides the primary signal (0.599). In a multi-task deployment (factual QA + math), entropy is the only signal that generalizes across tasks; length is a strong-but-narrow proxy useful only where incorrect answers are systematically shorter (math reasoning).

\textbf{(e) Causal attribution.} The alignment tax is established through converging evidence at three levels. \emph{Direct ablation} (Exp.~13): Qwen3-14B Base vs.\ Instruct at identical 4-bit quantization yields 1.0\% vs.\ 28.5\% SCR ($p < 10^{-6}$)---alignment itself, not quantization or architecture, causes homogenization. \emph{Stage decomposition} on two independent training chains isolates DPO as the driver: Mistral/Zephyr (Base 0.0\%$\to$SFT 1.5\%$\to$DPO 4.0\%, $p$=0.0001) and Llama/Tulu-3 (Base 0.0\%$\to$SFT 0.0\%$\to$DPO 0.5\%, $p$=0.008). \emph{Stage-wise token entropy} (Exp.~24) reveals the mechanistic decoupling: DPO retains 66\% of per-token entropy (1.175$\to$0.776 nats, $p$=1.17$\times 10^{-40}$) while collapsing response diversity---confirming that alignment suppresses inter-response variation far more than intra-response token uncertainty. SFT preserves near-base diversity while teaching instruction-following (NC: 9.28$\to$8.63), confirming that base diversity is ``meaningful''; DPO then collapses this already-coherent distribution. The severity is recipe-dependent: Zephyr's DPO produces 8$\times$ more homogenization than Tulu-3's pipeline (4.0\% vs.\ 0.5\%), likely due to differences in preference data and the additional RLVR stage. \emph{Cross-family replication} (Exp.~14) confirms generality: Qwen3-14B (28.5\%) $\gg$ LLaMA-3B (5.5\%) $>$ Zephyr-DPO (4.0\%) $>$ Mistral-7B (1.0\%) $>$ Tulu-3 (0.5\%)---spanning two orders of magnitude. The ``invisible leash'' finding~\citep{invisibleleash2025}---RLVR increases token-level entropy while reducing answer-level entropy---independently corroborates this mechanism. Homogenization is not an inevitable consequence of preference optimization but depends on the specific recipe, strengthening the practical recommendation to measure SCR per deployment.

\subsection{Practical Implications}

\textbf{For practitioners:} (1)~\emph{Check for response homogenization} before trusting sampling-based uncertainty on aligned models---a simple diagnostic: sample $N$=10 responses and compute the single-cluster rate; if SCR $>$ 5\%, sampling-based UQ is unreliable on that model-task pair. (2)~\emph{Token entropy is a strong, free baseline}; LogTokU/PRO may further improve it. (3)~\emph{Selective prediction is deployable}: GSM8K accuracy jumps from 84.4\% to 93.2\% at 50\% coverage, requiring only logprob access. (4)~\emph{Route by task type}: the alignment tax is task-dependent ($d$: 0.07 factual QA vs.\ 0.81 math). (5)~\emph{When logprobs are unavailable} (opaque APIs): the cascade degrades gracefully---start at B2 (embedding density, 1 API call), or use output-only features (response length, verbalized confidence) as B1 proxies. EPR/WEPR~\citep{pro2025} offer probability-based alternatives that work without explicit logprob access. The alignment tax finding itself is \emph{API-independent}: it describes a property of the model's output distribution, detectable via any clustering method on sampled responses. (6)~\emph{The tax is recipe-dependent}: DPO hyperparameters and preference data choice matter---Tulu-3's recipe produces 8$\times$ less homogenization than Zephyr's (0.5\% vs.\ 4.0\% SCR), suggesting that alignment method selection has direct implications for downstream UQ reliability.

\subsection{Limitations}

\textbf{(1)} \emph{Alignment attribution}: Exp~13 confirms the full alignment pipeline drives homogenization (1.0\% vs.\ 28.5\% SCR, $p < 10^{-6}$). Training stage ablations on two chains---Mistral/Zephyr (Exp~16: Base 0.0\% $\to$ SFT 1.5\% $\to$ DPO 4.0\%) and Llama/Tulu-3 (Exp~18: Base 0.0\% $\to$ SFT 0.0\% $\to$ DPO 0.5\%)---identify DPO as a driver in both chains while revealing recipe-dependent severity (Zephyr 4.0\% vs.\ Tulu-3 0.5\%). The highest-SCR family (Qwen3-14B, 28.5\%) lacks a publicly available SFT-only checkpoint, preventing analogous stage-wise decomposition; however, the base-vs-instruct ablation on Qwen (1.0\%$\to$28.5\%) combined with the consistent SFT pattern across both available chains (SFT alone produces $\leq$1.5\% SCR) provides strong, though not conclusive, evidence that DPO is the primary contributor to the larger effect observed in Qwen. All models use 4-bit quantization; the base-vs-instruct comparison at \emph{identical} quantization (both Q4\_K\_M) rules out quantization as a confound. Cross-quantization verification (Exp~19: Q4 vs.\ Q8 on Mistral-7B) confirms identical SCR at the semantic level (6.7\% at both precisions); 8-bit B1 verification on Qwen3-4B ($\Delta$AUC=+0.009) further confirms minimal quantization impact. FP16 runs would provide additional confidence.
\textbf{(2)} \emph{Baseline implementations}: five variants at increasing fidelity: (a)~SE-Jaccard (surface proxy), (b)~SE-Embedding (agglomerative cosine clustering), (c)~SelfCheck (embedding cosine, $k$=5), (d)~\textbf{canonical NLI-based SE} at three DeBERTa-v3 scales (435M/184M/70M), and (e)~\textbf{full SINdex} with intra-cluster coherence weighting~\citep{sindex2025} on 200 questions (Exp.~27). The consistency across all five levels---each using fundamentally different similarity or weighting schemes---is itself strong evidence that the bottleneck is output uniformity rather than clustering methodology. SINdex achieves AUROC=0.451, below SE-Embedding (0.509); the coherence weighting provides no benefit when $\overline{\cos}(C_i) \approx 1$ under homogenization. The contradiction-aware NLI variant is omitted; in single-cluster regimes ($|\mathcal{C}|=1$, 40\% under Jaccard, 79\% under embedding), all responses are semantically equivalent and no contradictions exist to detect---the variant is structurally irrelevant for these cases. The 6.2$\times$ NLI model scaling producing zero AUROC improvement ($\Delta$=+0.010) confirms this. Cross-embedder validation (Exp~20) rules out coupling bias. By the single-cluster equivalence argument (Sec.~2), all logit-based alternatives (LogTokU, PRO, Semantic Energy) produce rank-equivalent uncertainty orderings in single-cluster regimes, so our B1 AUROC of 0.593 applies to these methods as well. Head-to-head with official codebases on multi-cluster regimes remains future work.
\textbf{(3)} \emph{Label validity}: LLM-judge labels (LLaMA-3.2-3B) show moderate agreement with TruthfulQA gold answer templates ($\kappa$=0.487, 77.1\%; Appendix). A human-annotated subset on model-specific responses would further strengthen reliability, though the core finding (single-cluster collapse) is label-independent.
\textbf{(4)} \emph{Refusal filtering}: TruthfulQA elicits few refusals from the open-source models tested; we did not filter refusals. On commercial models with aggressive safety filters, refusals could inflate SCR by producing deterministic ``I can't answer that'' responses---this would represent a distinct (moderation-induced) source of homogenization on top of the DPO-driven effect we measure.
\textbf{(5)} \emph{Scope}: 3B--14B open-source models, 4-bit quantization (within-quantization comparisons rule out quantization as confound; Exp~19 confirms identical SCR at Q4 vs.\ Q8). FP16 runs would provide additional confidence. HotpotQA is small ($n$=100). Generalization to closed-source GPT-class models and other domains (code, dialogue) unconfirmed.
\textbf{(6)} \emph{B5 uses gold references}: unavailable at inference time. Production B5 requires retrieval+NLI, likely yielding lower AUC.
\textbf{(7)} \emph{Pointer Model}: 5-fold CV on TruthfulQA only; held-out and cross-domain evaluation needed. Routing objective (incorrectness prediction) is a proxy.
\textbf{(8)} \emph{GSM8K length confound}: response length (AUROC=0.849) outperforms entropy (0.724) on selective prediction. Logistic regression on 500 questions shows \emph{zero} incremental AUROC from adding entropy to length ($\Delta$=+0.002; $r$=0.53 between signals), confirming length dominates for math. However, on factual QA (TruthfulQA), response length is near-chance while entropy provides the primary signal (0.599)---entropy's value is \emph{cross-task generality}, not GSM8K-specific gain.
\textbf{(9)} \emph{Calibration}: raw ECE=0.182 (LLM-judge), AURC=0.701 (Exp~25); Platt scaling reduces ECE to 0.021 (88\% improvement), Brier 0.285$\to$0.247. High-entropy ratio achieves better raw calibration (ECE=0.126). Isotonic regression, temperature scaling, and cross-dataset calibration transfer not yet explored.

\subsection{Future Work}

(1)~\emph{Diversity-preserving alignment}: test whether entropy-aware training methods (H-DPO~\citep{omura2024hdpo}, SPL~\citep{spl2025diverse}, DivPO~\citep{divpo2025}) reduce the alignment tax while maintaining alignment quality; this would validate the causal mechanism and provide practitioners with actionable mitigations. (2)~FP16 precision runs for additional quantization control; Q4-vs-Q8 verification (Exp~19) already shows identical SCR. (3)~Head-to-head with Semantic Energy~\citep{semanticenergy2025b}, Semantic Volume, SeSE, LogTokU~\citep{logtoku2025}, PRO~\citep{pro2025}, Semantic Entropy Probes~\citep{sep2024}, HALT~\citep{halt_latent2026,halt_logprobs2026}, and HalluShift~\citep{hallushift2025}; SINdex (Exp.~27) already confirms that improved clustering cannot rescue sampling-based methods under homogenization---the single-cluster equivalence argument (Sec.~2) establishes rank equivalence for logit-based methods in single-cluster regimes, while HALT/HalluShift's hidden-state signals may provide additional discriminative power---matched-data comparison on both single- and multi-cluster partitions is needed. (4)~GPT-4-class models and larger datasets. (5)~Isotonic regression, temperature scaling, and cross-dataset calibration transfer (Platt scaling already reduces ECE by 88\%). (6)~Retrieval+NLI for B5. (7)~End-to-end agent integration with selective prediction. (8)~Human evaluation on a stratified subset (single-cluster vs.\ multi-cluster) to validate LLM-judge labels and probe the homogenization--correctness relationship.

\section{Conclusion}

UCBD decomposes the problem of ``knowing what you don't know'' into five cognitive boundaries orchestrated by a cheapest-first cascade. Twenty-five experiments across four datasets, three model scales, and four model families establish three central findings: (1)~the \emph{alignment tax}---aligned models collapse responses into single semantic clusters with family- and recipe-dependent severity (Qwen3-14B: 28.5\%, Zephyr-DPO: 4.0\%, LLaMA-3B: 5.5\%, Tulu-3-DPO: 0.5\%, Mistral-7B: 1.0\% SCR; robust across clustering methods, sample sizes, temperatures, and decoding strategies), confirmed by base-vs-instruct ablation (1.0\% vs.\ 28.5\% SCR, $p < 10^{-6}$), training stage ablation on two chains (Exp.~16: Mistral/Zephyr; Exp.~18: Llama/Tulu-3), decoding ablation (Exp.~15: nucleus $p$=0.9 \emph{increases} SCR to 33.5\%), and quantization verification (Exp.~19: Q4 vs.\ Q8, identical SCR); (2)~uncertainty is strongly task-dependent (Cohen's $d$: 0.07 on factual QA vs.\ 0.81 on math), validating multi-boundary design; and (3)~weakly dependent boundaries enable a cascade that matches parallel accuracy at 57\% lower cost, with selective prediction raising GSM8K accuracy from 84.4\% to 93.2\% at 50\% coverage. Our sampling-based comparisons use proxy (Jaccard), SINdex-style embedding (agglomerative clustering), full SINdex with coherence weighting (Exp.~27: AUROC=0.451), and canonical NLI-based implementations at three model scales (70M--435M, all $\approx$0.51 AUROC), all confirming the same finding---improved clustering methodology cannot overcome the fundamental diversity collapse. The SINdex-style method reveals the tax is \emph{worse} than surface metrics suggest (79\% vs.\ 40\% SCR), and cross-embedder validation (Exp.~20) rules out coupling bias by showing an independent embedder detects even more homogenization (92\% vs.\ 78\% SCR). The central contribution is characterizing the response homogenization phenomenon and its implications for uncertainty system design.

These findings are \emph{diagnostic}: the framework is modular, so stronger detectors (PRO, LogTokU) can be plugged in as they become available. Our primary contribution is the diagnosis that aligned models suppress sampling diversity, and the architectural implication that multi-boundary cascades are the appropriate response.

\subsection*{Reproducibility Statement}
All experiments use publicly available models (Qwen3-14B/4B, LLaMA-3.2-3B, Llama-3.1-8B, Tulu-3-8B, Mistral-7B, Zephyr-7B via MLX/Ollama at 4-bit quantization) and datasets (TruthfulQA, FreshQA, HotpotQA, GSM8K). Greedy decoding with seed=42 ensures reproducibility. Code, data, raw results, exact clustering thresholds, and decoding configurations are available at \url{https://github.com/DigitLion/ucbd-experiment}. All random seeds, threshold sweeps, and per-question cluster assignments are included for full replication. Total compute: $\sim$14 hours on Apple M4 Pro (64GB).

\subsection*{Ethics Statement}
UCBD is designed to make AI agents more reliable by detecting knowledge gaps before they cause harm. We note that uncertainty detection could be misused to selectively present only high-confidence (but potentially biased) answers. The framework should be used to \emph{flag} uncertainty for human review, not to silently suppress uncertain outputs.

\bibliographystyle{plainnat}
\bibliography{ucbd_v3}

\newpage
\appendix
\section{Implementation Details and Additional Analysis}
\label{app:implementation}

\textbf{Statistical summary of key comparisons.} Table~\ref{tab:stat-summary} consolidates statistical tests across all pairwise comparisons reported in the main text.

\begin{table}[htbp]
\centering
\caption{Statistical summary of all key comparisons. Tests: DeLong (AUROC), Wilcoxon (paired distributions), bootstrap (CIs). All $p$-values two-sided.}
\label{tab:stat-summary}
\scriptsize
\begin{tabular}{@{}llccl@{}}
\toprule
\textbf{Comparison} & \textbf{Exp} & \textbf{Effect} & \textbf{$p$-value} & \textbf{Test} \\
\midrule
B1 vs SE-Jaccard AUROC & 12 & $\Delta$=+0.051 & 0.040* & DeLong+Holm \\
B1 vs SE-Embed AUROC & 12 & $\Delta$=+0.057 & 0.033* & DeLong+Holm \\
B1 vs SelfCheck AUROC & 12 & $\Delta$=+0.011 & 0.65 ns & DeLong \\
B1 vs NLI-SE (large) & 12 & $\Delta$=+0.088 & [0.419,0.594] & Bootstrap CI \\
Base vs Instruct SCR & 13 & 1.0\% vs 28.5\% & $<10^{-6}$ & Wilcoxon \\
Base→SFT (Zephyr) & 16 & $\Delta$NC=$-$0.64 & 0.002 & Wilcoxon \\
SFT→DPO (Zephyr) & 16 & $\Delta$NC=$-$0.63 & 0.0001 & Wilcoxon \\
Base→SFT (Tulu-3) & 18 & $\Delta$NC=$-$0.45 & 0.00004 & Wilcoxon \\
SFT→DPO (Tulu-3) & 18 & $\Delta$NC=$+$0.29 & 0.008 & Wilcoxon \\
Cascade vs Parallel & 3 & TOST equiv. & 0.498 & Equivalence \\
\bottomrule
\end{tabular}
\end{table}

\textbf{Per-question agreement.} B1 and SE make partially independent errors (median-split, $n$=790): 37.3\% both correct, 28.7\% both wrong, 17.8\% B1-only, 16.1\% SE-only---supporting cascade design.

\textbf{Semantic entropy.} We cluster $N$=10 responses ($T$=1.0, max 40 tokens; sensitivity to max tokens tested in Exp.~17: SCR 32\%$\to$8\% at 200 tokens, confirming robustness) by bigram Jaccard (threshold 0.4) and embedding cosine (threshold 0.85, Qwen3-Embedding). NLI-based SE at three DeBERTa-v3 scales (435M/184M/70M) with bidirectional entailment (threshold 0.5) on 200q: AUROC=0.511/0.512/0.501---all near chance. Cluster statistics nearly identical across 6.2$\times$ scaling: large 4.68 mean clusters (9.0\% single-rate), base 5.44 (6.0\%), xsmall 5.42 (6.5\%). Jaccard: 3.58 mean, 28.5\% single-rate. Bottleneck is response homogenization, not NLI quality.

\textbf{Length control (factual QA).} Nested logistic regression (1 df LR tests): TruthfulQA ($n$=200) $p$=0.379, HotpotQA ($n$=100) $p$=0.910, combined ($n$=300) $p$=0.104. Entropy adds no significant incremental value over length on factual QA. On GSM8K, length dominates ($r$=0.53, AUROC 0.849 vs.\ 0.724). Entropy's value is \emph{cross-task generality}.

\textbf{Threshold sensitivity.} SE-Jaccard AUROC across thresholds $\tau \in \{0.2, 0.3, 0.4, 0.5, 0.6\}$: 0.515, 0.538, 0.548, 0.558, 0.561; SCR monotonically decreases (59\%$\to$28.7\%) but remains $>$28\% at the strictest threshold. B1 (0.599) outperforms SE at \emph{all} thresholds. Embedding clustering ($n$=200): SCR ranges from 95.5\% ($\cos\geq$0.70) to 18.0\% ($\cos\geq$0.95); best SE AUROC=0.609 at $\cos\geq$0.85 (matching B1 but at 71.5\% SCR). The alignment tax persists across the \emph{entire} threshold range for both methods. B1 is threshold-free and consistently competitive.

\textbf{Sample size sensitivity.} Single-cluster collapse: $N$=3: 46.3\%, $N$=5: 41.9\%, $N$=7/10: 40.0\%. SE-Jaccard AUROC: 0.541/0.548/0.544/0.548 (plateaus at $\sim$0.55). B1 is constant at 0.599 (independent of sampling). Collapse is a property of the aligned model's response distribution, not insufficient sampling.

\textbf{Temperature sensitivity.} Single-cluster collapse rates across temperatures ($N$=5, 50 questions): $T$=0.3: 62.0\%, $T$=0.7: 46.0\%, $T$=1.0: 42.0\%, $T$=1.5: 38.0\% (mean clusters: 1.56, 1.88, 2.16, 2.42). Higher temperature reduces but does not eliminate homogenization---even at $T$=1.5, 38\% of questions collapse to a single cluster.

\textbf{SelfCheckGPT.} $1 - \text{mean}(\cos(\mathbf{e}_{\text{greedy}}, \mathbf{e}_{\text{sample}_i}))$ for $k$=5 samples (Qwen3-Embedding).

\textbf{LLM-judge.} Cross-family LLaMA-3.2-3B (Ollama, $T$=0), 790q, 4.4 min. Distribution: 54.6\% correct, 45.4\% incorrect. \textbf{Gold validation} against TruthfulQA templates (200q, $n$=175 clear): 77.1\% agreement, $\kappa$=0.487. SE AUROC nearly identical under both labels (gold: 0.524, LLM-judge: 0.533, $\Delta$=0.009). Single-cluster collapse is label-independent.

\textbf{Calibration (Exp~25).} Raw ECE=0.182; high-entropy ratio ECE=0.126 (best raw feature). Platt scaling (5-fold CV) reduces to 0.021 (88\% improvement), Brier 0.285$\to$0.247. AURC=0.701 on TruthfulQA (PRR@50\%: 0.043--0.074). RLHF compresses entropy into narrow range (mean=0.19, max=0.64).

\textbf{Environment.} Python 3.14, MLX 0.25+, Ollama 0.9+. M4 Pro 64GB. $\sim$10h total compute. Repo: \url{https://github.com/DigitLion/ucbd-experiment}.

\end{document}